\documentclass[sigconf]{acmart}

\setcopyright{acmlicensed}
\copyrightyear{2025}
\acmYear{2025}
\acmDOI{XXXXXXX.XXXXXXX}
\acmConference[Conference acronym 'XX]{Conference Title}{June 2025}{Woodstock, NY}
\acmISBN{978-1-4503-XXXX-X/2018/06}

\usepackage{microtype}
\usepackage{graphicx}
\usepackage{subcaption}
\usepackage{booktabs} 
\usepackage{xcolor,colortbl}
\usepackage{multirow}
\usepackage{algorithm}
\usepackage{algpseudocode}
\usepackage{paralist}
\usepackage{enumitem}
\usepackage{tabularx}
\usepackage{soul}
\usepackage{amsmath}
\usepackage{amssymb}
\usepackage{mathtools}
\usepackage{amsthm}
\usepackage[capitalize,noabbrev]{cleveref}
\usepackage[textsize=tiny]{todonotes}
\usepackage{makecell}

\definecolor{low1}{RGB}{255, 204, 204}
\definecolor{low3}{RGB}{255, 255, 204}
\definecolor{low2}{RGB}{204, 255, 204}

\theoremstyle{plain}
\newtheorem{theorem}{Theorem}[section]

\theoremstyle{definition}

\theoremstyle{remark}

\usepackage[normalem]{ulem}

\title{Pre-training Epidemic Time Series Forecasters with Compartmental Prototypes}

\author{Zewen Liu}
\affiliation{%
  \institution{Emory University}
  \department{Department of Computer Science}
  \city{Atlanta}
  \state{GA}
  \country{USA}
}
\email{zewen.liu@emory.edu}

\author{Juntong Ni}
\affiliation{%
  \institution{Emory University}
  \department{Department of Computer Science}
  \city{Atlanta}
  \state{GA}
  \country{USA}
}
\email{juntong.ni@emory.edu}

\author{Bohan Wang}
\affiliation{%
  \institution{Emory University}
  \department{Department of Computer Science}
  \city{Atlanta}
  \state{GA}
  \country{USA}
}
\email{bohan.wang@emory.edu}

\author{Max S. Y. Lau}
\affiliation{%
  \institution{Emory University}
  \department{Rollins School of Public Health}
  \city{Atlanta}
  \state{GA}
  \country{USA}
}
\email{msy.lau@emory.edu}

\author{Wei Jin}
\affiliation{%
  \institution{Emory University}
  \department{Department of Computer Science}
  \city{Atlanta}
  \state{GA}
  \country{USA}
}
\email{wei.jin@emory.edu}

\begin{document}

\begin{abstract}
Accurate epidemic forecasting is crucial for outbreak preparedness, but existing data-driven models are often brittle. Typically trained on a single pathogen, they struggle with data scarcity during new outbreaks and fail under distribution shifts caused by viral evolution or interventions. However, decades of surveillance data and the design of various compartmental models from diverse diseases offer an untapped source of transferable knowledge. 
To leverage the collective lessons from history, we propose CAPE, the first open-source pre-trained model for epidemic forecasting.
Unlike existing time series foundation models that overlook epidemiological challenges, CAPE models epidemic dynamics as mixtures of latent compartmental population states, termed \textit{compartmental prototypes}. It models a flexible dictionary of compartment prototypes directly from a large collection of simulation data, enabling each outbreak to be expressed as a time-varying mixture that links observed infections to latent population states. To promote robust generalization, CAPE adopts the next-token-prediction paradigm during pre-training with lightweight epidemic-aware regularization that aligns the learned prototypes with epidemiological semantics. On a comprehensive benchmark spanning 17 diseases, CAPE significantly outperforms strong baselines with zero-shot forecasting. This work represents a principled step toward pre-trained epidemic models that are both transferable and epidemiologically grounded. We provide our code in: \url{https://github.com/nuuuh/CAPE}.
\end{abstract}

\maketitle

\section{Introduction}
Infectious disease outbreaks pose a persistent threat to global public health and economic stability~\citep{nicola2020socio}. Effective outbreak management relies on accurate epidemic forecasting—the prediction of future cases, hospitalizations, and other critical metrics~\citep{liu2024review, wan2024epidemiology, adhikari2019epideep}. A wide range of models have been developed to provide these crucial forecasts, which generally fall into two categories. Knowledge-driven mechanistic models, such as the classic Susceptible-Infected-Recovered (SIR)~\citep{cooper2020sir} approach, are grounded in epidemiological principles; they divide the population into \textit{compartments} that represent distinct \textit{population states} (e.g., susceptible, infectious, recovered) and use differential equations to explicitly model flows among these states. In contrast, modern data-driven methods like LSTMs~\citep{shahid2020predictions} learn complex patterns directly from historical data, offering greater flexibility without imposing a predefined structure of dynamics.

However, these data-driven forecasters are often trained for a single pathogen. This narrow scope makes them brittle: they face acute data scarcity during the critical early stages of a novel outbreak, and they fail under distribution shifts induced by viral evolution. While training across diverse pathogens and geographies proves to enhance downstream forecasting tasks~\cite {kamarthi2023pems}, real-world data often lacks of observation of hidden population groups (e.g., susceptible, exposed, vaccinated), which are essential for disclosing the underlying disease dynamics. Nevertheless, with decades of studies across various diseases, epidemiologists have uncovered diverse disease dynamics through mathematical modeling with extensive prior knowledge, which are often formulated as coupled ordinary differential equations, capable of simulating disease outbreaks. Motivated by the success of physics-informed neural networks in epidemic forecasting~\cite{wan2024epidemiology, rodriguez2023einns}, which make use of such prior knowledge, and large pre-trained models in language, vision, and time-series domains~\citep{zhao2023survey}, we ask: \textit{Can we build a pre-trained epidemic forecaster that learns from the collective dynamics of infectious diseases to generalize effectively across real-world outbreaks?}

Simply applying a general time series foundation model~\citep{liang2024foundation} or training an epidemic-informed neural network is insufficient, as it overlooks core epidemiological challenges for generalization: (1) \textit{Structural heterogeneity}: Pathogens follow different effective compartmental progressions (e.g., SIR vs. SEIR~\citep{he2020seir}), so a single fixed mechanism cannot transfer broadly across diseases and regions. (2) \textit{Hidden population states}: Surveillance data records only reported infections, while important states such as exposure, susceptibility, and immunity are not directly observed. These properties demand both large-scale and diverse simulations of diseases to capture various dynamics and a powerful epidemic pre-trained model structure that can adapt to different pathogens and provide forecasts under different hidden population groups.

\textbf{Our Solution.} \textit{First}, we introduce \textbf{EpiRecipe}, a pipeline for building compartmental models and conducting large-scale simulations of diverse disease dynamics.
\textit{Second}, with the produced supervision data from EpiRecipe and real-world historical disease data, we introduce a novel pre-training-based epidemic model: CAPE (\underline{C}omp\underline{A}rtment \underline{P}re-training for \underline{E}pidemics), which learns epidemic dynamics as a mixture of latent population groups, termed \textbf{compartmental prototypes}. To address structural heterogeneity and hidden states, rather than relying on a rigid, pre-defined structure, CAPE learns a flexible dictionary of prototypes directly from data via next-token prediction. This allows the model to transfer to downstream forecasting tasks by dynamically composing these prototypes via compartmental masking. Our contributions include:

\begin{compactenum}[(1)]

\item \textbf{Large Scale Simulation Pipeline:} To enlarge the pre-training corpus, we propose \textbf{EpiRecipe}, a comprehensive simulation pipeline that maintains a catalog of 12 epidemiological compartments and over 20 transition dynamics to generate millions of structurally valid, diverse disease simulations

\item  \textbf{Pre-training framework for epidemic time series forecasting:} We introduce the first open-source pre-training framework\footnote{\url{https://github.com/nuuuh/CAPE}} for epidemic forecasting. It learns latent compartmental prototypes directly from disease dynamics and provides uncertainty quantification by conducting stochastic structural inference via compartmental masking.

\item \textbf{Comprehensive evaluation benchmark and state-of-the-art performance:} We assemble a comprehensive evaluation pipeline spanning 17 diverse diseases in the US under an online forecasting setting. Even in a zero-shot setting, CAPE exhibits the best MSE and MAE against baselines and achieves a stronger performance compared to time series foundation models. 

\item \textbf{In-depth analysis:} We conduct extensive analyses to uncover how pre-training improves representation learning. We provide the first evidence of \textit{epidemic scaling laws} on simulation data and visualize how the model captures important epidemic properties like peak timing without explicit supervision.

\end{compactenum}

\section{Related Work and Problem Definition}

\textbf{Epidemic Forecasting Models.}
Traditionally, epidemic forecasting employs models like ARIMA~\citep{sahai2020arima}, SEIR~\citep{he2020seir}, and VAR~\citep{shang2021regional}. ARIMA predicts infections by analyzing past data and errors, SEIR models population transitions using differential equations, and VAR captures linear inter-dependencies by modeling each variable based on past values. Recently, deep learning models, categorized into RNN-based, MLP-based, and transformer-based, have surpassed these methods. RNN-based models like LSTM~\citep{wang2020time}, GRU~\citep{natarajan2023outbreak}, and more epidemic-specific models like EpiDeep~\cite{adhikari2019epideep} and EINNs~\cite{rodriguez2023einns} use gating mechanisms to manage information flow. MLP-based models use linear layers~\citep{zeng2023transformers} or multi-layer perceptrons~\citep{borghi2021covid,madden2024deep} for efficient data-to-prediction mapping and physics-informed distillation~\cite{wang2021deep}. Transformer-based models~\citep{wu2021autoformer, zhou2021informer, zhou2022fedformer} apply self-attention to encode time series and generate predictions via a decoder. However, these models are limited as they typically utilize data from only one type of disease without considering valuable insights from diverse disease datasets.

\noindent\textbf{Pre-trained Time Series Models.}
To enable few-shot or zero-shot capabilities, transformer-based models often employ pre-training on large datasets, which typically use masked data reconstruction~\citep{zerveas2021transformer, rasul2023lag} or promote alignment across different contexts~\citep{fraikin2023t, zhang2022self, yue2022ts2vec}. For example, PatchTST~\citep{nie2022time} segments time series into patches, masks some, and reconstructs the masked segments. Larger foundational models like MOMENT~\citep{goswami2024moment} and Chronos~\cite{ansari2024chronos} aim to excel in multiple tasks (e.g., forecasting, imputation, classification) and prove useful in epidemic forecasting~\cite{panja2025zero, kalahasti2025foundation, dey2024we}, but training them requires substantial data and computational resources. In epidemic contexts, Kamarthi et al.~\citep{kamarthi2023pems} pre-train a model on various diseases, improving downstream performance and highlighting pre-training's potential in epidemic forecasting. However, the complete implementation is not publicly available. Moreover, existing approaches overlook hidden compartmental influence and zero-shot ability in epidemic forecasting and lack a deep analysis of how pre-training materials impact downstream performance. In this study, we introduce latent compartment modeling and conduct a thorough analysis of these questions.

\noindent\underline{\textbf{Problem Definition.}}
Given historical observations $\mathbf{x} \in \mathbb{R}^{T}$, the goal is to forecast future infections $\mathbf{y} \in \mathbb{R}^{H}$, where $T$ and $H$ denote the lookback window and forecast horizon, respectively. We employ a pre-training framework on a hybrid corpus $\mathcal{D}_{\text{pre}} = \mathcal{D}_{\text{syn}} \cup \mathcal{D}_{\text{real}}$, comprising synthetic samples from EpiRecipe and historical real-world data. Adopting a \textit{next-token-prediction} paradigm, we define a patching operator $\mathcal{P}: \mathbb{R}^T \to \mathbb{R}^{N \times P}$ that segments $\mathbf{x}$ into a sequence of $N$ tokens $\mathbf{X'} = (\mathbf{x'}_1, \dots, \mathbf{x'}_N)$, where each $\mathbf{x'}_i \in \mathbb{R}^P$ is a patch of length $P$. The model $f_\theta$ is trained to autoregressively predict the next token $\mathbf{x'}_{t+1}$ and latent epidemic states $\mathbf{z}_{t+1}$ (e.g., compartmental dynamics, $R_t$) conditioned on the context $\mathbf{X'}_{1:t}$. The optimization objective is:
\begin{equation}
    \min_\theta \mathbb{E}_{\mathbf{x} \sim \mathcal{D}_{\text{pre}}} \left[ \sum\nolimits_{t=1}^{N-1} \left( \mathcal{L}_{\text{rec}}( \hat{\mathbf{x'}}_{t+1}, \mathbf{x'}_{t+1} ) + \lambda \mathcal{L}_{\text{epi}}( \hat{\mathbf{z}}_{t+1}, \mathbf{z}_{t+1} ) \right) \right]
\end{equation}
where $\hat{\mathbf{x'}}_{t+1}, \hat{\mathbf{z}}_{t+1} = f_\theta(\mathbf{X'}_{1:t})$ are the predicted token and latent states, and $\lambda$ balances the reconstruction loss $\mathcal{L}_{\text{rec}}$ with the epidemic consistency loss $\mathcal{L}_{\text{epi}}$.

\section{Proposed Method}

Our pre-training framework aims to overcome the core challenges of \textit{structural heterogeneity} and \textit{hidden population states} inherent in epidemic forecasting. We address these issues through two main contributions: (1) a comprehensive pipeline for producing random systems of disease dynamics (section ~\ref{sec: epirecipe}), and (2) a flexible model architecture that learns latent compartmental prototypes directly from simulation data, with epidemic-aware pre-training objectives that guide the model to learn robust, generalizable representations (section ~\ref{sec: cape}). In addition, derived naturally from the model structure, we also introduce a stochastic structural inference for both effective adaptation to downstream tasks and uncertainty estimation (section ~\ref{sec: train_infer}).

\begin{figure}
    \centering
    \includegraphics[width=0.9\linewidth]{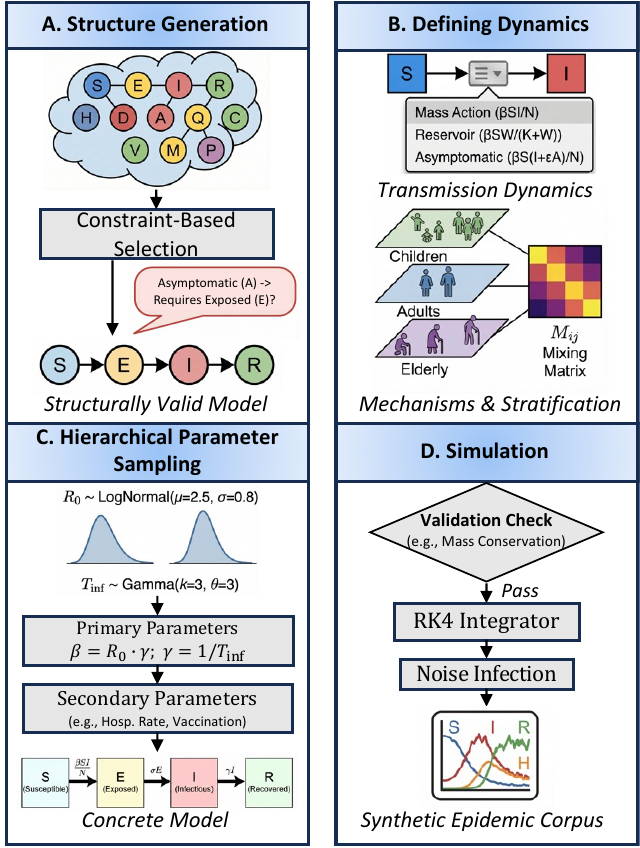}
    \vskip -1.1em
    \caption{Pipeline of EpiRecipe.}
    \label{fig:epirecipe}
\end{figure}

\subsection{EpiRecipe}
\label{sec: epirecipe}

A fundamental challenge in pre-training epidemic forecasting models is the scarcity and heterogeneity of real-world outbreak data. To address this, we develop \textbf{EpiRecipe}, a constraint-based simulation pipeline for generating diverse, epidemiologically valid compartmental models and their corresponding synthetic time series. This augments our pre-training corpus by exposing the model to a wide range of disease dynamics. The overall pipeline is shown in Figure~\ref{fig:epirecipe} and more details are provided in Appendix~\ref{app:epirecipe}.

\noindent\textbf{Constraint-Based Model Generation.}
To ensure coverage of diverse compartment states and resolve the challenge of structural heterogeneity behind disease outbreaks, EpiRecipe maintains a catalog of 12 epidemiological compartments~\cite{brauer2008compartmental}, representing diverse population groups (e.g., infectious individuals (\textbf{I})):
\begin{equation}
    \mathcal{C} = \{\textbf{S}, \textbf{E}, \textbf{I}, \textbf{R}, \textbf{H}, \textbf{V}, \textbf{Q}, \textbf{D}, \textbf{P}, \textbf{W}, \textbf{A}, \textbf{C}\},
\end{equation}
along with their dependency constraints encoded as requirements, defined as a mapping $f_\text{req}: \mathcal{C} \to \mathcal{P}(\mathcal{C})$, where each compartment maps to a subset of required prerequisite compartments $\mathcal{P}(\mathcal{C})$. For instance, the Infectious (\textbf{I}) compartment requires the Susceptible (\textbf{S}) compartment to exist, i.e., $\textbf{E} \in f_\text{req}(\textbf{A})$~\cite{anggriani2022mathematical}.

\noindent\textbf{Automatic Transition Construction.}
Even with the same set of compartments, different diseases can have diverse transition dynamics. Therefore, once compartments are selected, EpiRecipe automatically constructs a transition network $\mathcal{T}$ by randomly selecting from predefined mechanistic variants, where each transition from compartment $i$ to $j$ is governed by a flow function $f_{ij}$. For example, infection dynamics can follow standard mass-action $f_{\textbf{S}\to\textbf{E}} = \beta SI/N$~\cite{kolokolnikov2021law}. The pipeline selects appropriate variants of flow functions based on the available compartments, and the complete system evolves according to:
\begin{equation}
\frac{dx_c}{dt} = \underbrace{\sum_{i \to c} f_{ic}(\mathbf{x}; \theta)}_{\text{inflow}} - \underbrace{\sum_{c \to j} f_{cj}(\mathbf{x}; \theta)}_{\text{outflow}},
\label{eq:ode_system}
\end{equation}
where $\mathbf{x}$ is the state vector representing the population in each compartment, and the sums represent all incoming and outgoing transitions for compartment $c$.


\noindent\textbf{Population Stratification.}
Real epidemics exhibit heterogeneous dynamics across demographic groups. To capture this, EpiRecipe partitions the population into $G$ groups (e.g., children, adults, elderly), following the implementation in CovidSim~\cite{wilson2020modelling}. Each group $g$ is a population fraction and has its own group-specific properties like transmission rate $\beta_g(t)$. Inter-group transmission is governed by a mixing matrix $\mathbf{M} \in \mathbb{R}^{G \times G}$ with predefined patterns (e.g., homogeneous or assortative). The stratified infection dynamics become:
$f^{(g)}_{\textbf{S}\to\textbf{I}} = \beta_g \frac{S_g}{N_g} \sum_{h=1}^{G} M_{gh} \iota_h I_h$,
where $S_g$, $I_g$, and $N_g$ denote the susceptible, infectious, and total population in group $g$, respectively. This produces epidemic curves with realistic subpopulation dynamics and complex aggregate patterns. Specific sampling distributions for all multipliers are provided in Appendix~\ref{app:epirecipe}.

\noindent\textbf{Hierarchical Parameter Sampling.}
To ensure epidemiological realism, parameters are sampled hierarchically in three stages. \textit{First}, primary epidemiological quantities, e.g., the basic reproduction number $R_0$, are sampled from informed prior distributions. \textit{Second}, derived parameters are computed deterministically to maintain consistency, e.g., the baseline transmission rate $\beta_0 = R_0 \cdot \gamma$. Seasonal forcing is optionally applied via:
\begin{equation}
\beta(t) = \beta_0 \left[1 + \varepsilon \cos\left(\frac{2\pi t}{T_{\text{period}}} + \phi\right)\right],
\label{eq:seasonal}
\end{equation}
where $\varepsilon$, $T_{\text{period}}$, and $\phi$ control the amplitude, period, and phase of seasonal variation, respectively. \textit{Lastly}, secondary parameters required by the selected transitions (e.g., hospitalization and vaccination rates) are sampled from predefined ranges. All sampling distributions and ranges are provided in Appendix~\ref{app:epirecipe}.

\noindent\textbf{Validation and Simulation.}
Each generated model undergoes validation to ensure: (1) population conservation across living compartments, (2) no dead-end states, i.e., every non-terminal compartment has at least one outflow transition, and (3) existence of a valid infectious pathway from \textbf{S} to \textbf{I} in the transition graph. Valid models are then integrated using a fourth-order Runge-Kutta (RK4) solver, and Gaussian observation noise is injected to mimic real-world reporting imperfections:
\begin{equation}
\{\mathbf{x}(t)\}_{t=0}^{T} = \textsc{RK4}\left(\frac{d\mathbf{x}}{dt}, \mathbf{x}_0, T\right), \quad \tilde{\mathbf{x}}(t) = \max\left(\mathbf{x}(t) + \boldsymbol{\eta}(t), \mathbf{0}\right),
\label{eq:simulation}
\end{equation}
where $\boldsymbol{\eta}(t) \sim \mathcal{N}(\mathbf{0}, \lambda^2 \cdot \text{diag}(\text{Var}(\mathbf{x})))$ and $\lambda$ controls the noise level. Through this process, EpiRecipe generates millions of samples with varied compartmental structures, transition mechanisms, and parameter regimes, providing the diverse pre-training corpus essential for learning generalizable patterns.

\begin{figure*}
        \centering
        \includegraphics[width=\textwidth]{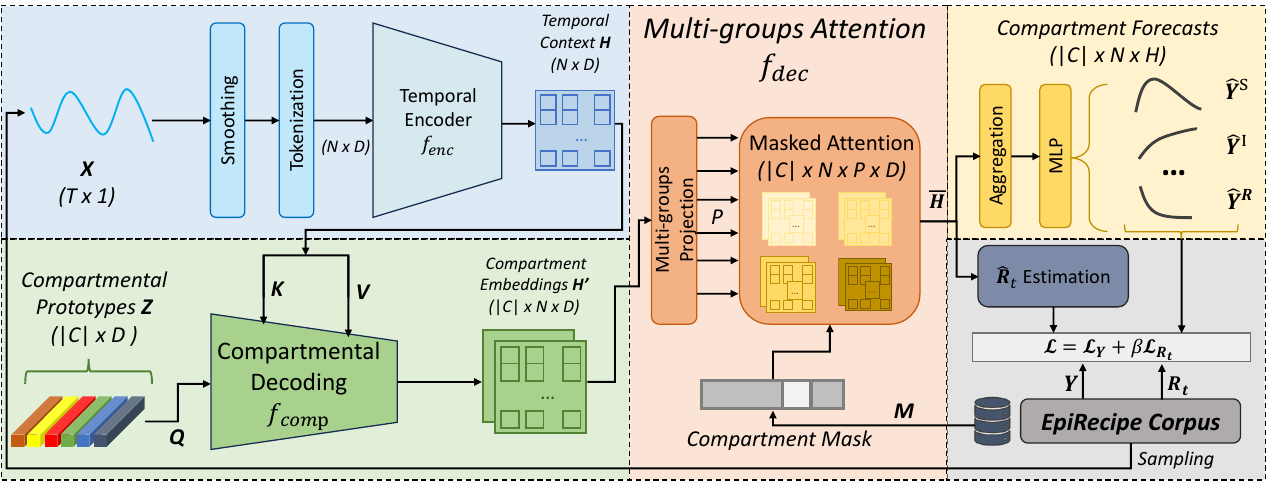}
        \vskip -0.5em
        \caption{CAPE model structure and pre-training pipeline with EpiRecipe.}
        \label{fig:CAPE}
        \vskip -1em
\end{figure*}

\vspace{-0.2em}
\subsection{Model Structure for Epidemic Pre-training}
\label{sec: model}

\label{sec: cape}

Similar to the current large language models and foundation models in time series~\cite{ansari2024chronos}, CAPE conducts next-token-prediction in an autoregressive way. CAPE is composed of three major components: (1) Autoregressive temporal encoder $f_{enc}$, (2) Compartmental decoding $f_{comp}$, and (3) Masked multi-groups attention $f_{dec}$. 

\noindent\textbf{Autoregressive Temporal Encoder ($f_{enc}$).} To effectively capture the evolving momentum and latent inertia of an outbreak, it is important to distill raw, often non-stationary infection trends into a high-dimensional temporal context that preserves both local fluctuations and global epidemiological trajectories. Given an observed infection trajectory $\mathbf{x} \in \mathbb{R}^{T \times 1}$, \textit{smoothing}~\cite{chung2020gaussian} and \textit{patching}~\cite{nie2022time} is performed to formulate the denoised token sequence $\mathbf{x'} \in \mathbb{R}^{N \times t}$, where $N$ is the number of tokens and $t=T/N$ is the token size. Then, an autoregressive model (e.g., LSTM, TCN, or Transformer) is applied to encode temporal information, producing embeddings for each patch $\mathbf{h_{i+1}}=f_{enc}(\mathbf{x'_0}, ..., \mathbf{x'_i}) \in \mathbb{R}^{D}$, where $i$ denotes the token index and $D$ is the embedding size. This encoding formulates the temporal context of past infections $\mathbf{H} \in \mathbb{R}^{N \times D}$.

\noindent\textbf{Compartmental Decoding ($f_{comp}$).}
Decoding directly from the univariate context is insufficient, as it not only ignores the provided supervision signals from different compartments but also the influence of hidden compartments on the forecast of future infections. Therefore, we propose to decode compartmental context in the latent space with both temporal context $\mathbf{H}$ and a codebook~\cite{zhou2022towards} of learnable compartmental prototypes, denoted as $\mathbf{Z} \in \mathbb{R}^{|C| \times D}$, where $C$ is a set of compartments that match with the ones in EpiRecipe. Finally, we use cross-attention to produce the contextualized compartment embeddings as:
\begin{equation}
\mathbf{H}' = f_{comp}(\mathbf{H}, \mathbf{Z}) = \text{softmax}\left({\mathbf{Z}\mathbf{H}^\top}/{\sqrt{D}}\right) \quad \mathbf{H} \in \mathbb{R}^{|C| \times N \times D}.
\label{eq:crossattn}
\end{equation}
Eventually, the prediction of the next token is decoded from the corresponding compartment embedding.

\noindent\textbf{Masked Multi-groups Attention ($f_{dec}$).} 
The observed dynamics of a compartment are often the result of complex, unobserved interactions among multiple latent population groups (e.g., age, occupation, or mobility-based strata) with differing contact rates and vulnerabilities. To bridge the gap between these hidden population states and visible compartmental outcomes, we propose \textit{Masked Multi-groups Attention}. By implicitly modeling these heterogeneous group dynamics through the regularization of the effective reproduction number $R_t$~\cite{lim2020interpretation}, where $R_t > 1$ signals expansion and $R_t < 1$ signals decline of infections, we enforce an epidemiologically coherent latent space that ensures the model's predictions align with the physical laws of disease transmission. Overall, this process produces predictions $\mathbf{Y}=f_{dec}(\mathbf{H}', \mathbf{M})$ with a compartment mask $\mathbf{M}$, as detailed below.

\textit{First}, each compartment $\mathbf{H}_{C_i}'$ is projected to $P$ multiple groups via linear projections $\mathbf{\hat{H}}_{C_i,j} = \mathbf{W}_j\mathbf{H}_{C_i}'$, each corresponding to a population group. Such an idea is similar to the linear projections in computing query, key, and value in self-attention and the mappings in multi-head attention~\cite{vaswani2017attention}, which enhances the expressiveness of the model. 
\textit{Second}, giving the mask $\mathbf{M}$, the masked self-attention is applied to model the interactions among compartments of interest in the latent space, producing $\mathbf{\Bar{H}}=\text{MaskedAttention}(\mathbf{\hat{H}}, \mathbf{M})$.
\textit{Third}, with the attention scores indicating the strength of contributions/interactions, we regularize the attention scores within the infectious group to match the actual dynamics. To bridge the gap between these latent attention patterns and $R_t$, we map the learned attention matrix to the components of the Next-Generation Matrix~\cite{diekmann2010construction}, where off-diagonal entries correspond to cross-group transmission rates (matrix $\mathbf{F}$) and diagonal entries reflect within-group stability (matrix $\mathbf{V}$).
Consequently, Theorem~\ref{thm:rt_bounds} provides the spectral bound for estimating $R_t$ (Proof in Appendix~\ref{Appendix: theory}).

\begin{theorem}[Bounds for $R_t$]
\label{thm:rt_bounds}
Let $\mathbf{F}, \mathbf{V} \in \mathbb{R}^{n\times n}$ with $\mathbf{V}$ invertible, and define $R_t \equiv \rho(\mathbf{F}\mathbf{V}^{-1})$, where $\rho(\cdot)$ is the spectral radius. Then:
\begin{equation}
\frac{\sigma_{\min}(\mathbf{F})}{\sigma_{\max}(\mathbf{V})} \;\leq\; R_t \;\leq\; \frac{\sigma_{\max}(\mathbf{F})}{\sigma_{\min}(\mathbf{V})},
\label{eq:rt_bounds}
\end{equation}
where $\sigma_{\max}(\cdot)$ and $\sigma_{\min}(\cdot)$ denote the maximal and minimal singular values, respectively.
\end{theorem}

Based on Theorem~\ref{thm:rt_bounds}, we derive a differentiable estimation of $\hat{R}_t$ from the attention scores via Algorithm~\ref{alg:rt_estimation}, and match the estimation with ground truth using $\mathcal{L}_{R_t}=\text{MSE}(\hat{R}_t, R_t)$. Lastly, for each compartment $C_i$, the latent groups are aggregated and projected to the next token via $\mathbf{Y}_i=\text{MLP}(\sum_{j=1}^{P}\Bar{H}_{C_i,j})$, where $\mathbf{Y}_i$ is the future trajectory of compartment $C_i$.

\begin{algorithm}[t]
\caption{$R(t)$ Estimation from Attention Patterns}
\label{alg:rt_estimation}
\begin{algorithmic}[1]
\Require Attention scores $\mathbf{A}$, number of views $P$
\Ensure $R(t)$ time series $\{\hat{R}_1, \hat{R}_2, \ldots, \hat{R}_T\}$

\For{$t = 1$ to $T$}
    \State Extract I-to-I attention submatrix: $\mathbf{A}_t^{(I)} \in \mathbb{R}^{P \times P}$
    
    \State \textbf{Construct F matrix:}
    \State \quad $\mathbf{F} \gets \mathbf{A}_t^{(I)}$, \quad $F_{ii} \gets 0$ \Comment{cross-group infections}
    
    \State \textbf{Construct V matrix:}
    \State \quad $\gamma_i \gets 1 / (A^{(I)}_{t,ii} + \epsilon)$ \Comment{within-group transitions}
    \State \quad $\mathbf{V} \gets \text{diag}(\gamma_1, \ldots, \gamma_P)$
    
    \State \textbf{Compute spectral bounds (Theorem~\ref{thm:rt_bounds}):}
    \State \quad $R_t^{\text{lower}} \gets \sigma_{\min}(\mathbf{F}) \,/\, (\sigma_{\max}(\mathbf{V}) + \epsilon)$
    \State \quad $R_t^{\text{upper}} \gets \sigma_{\max}(\mathbf{F}) \,/\, (\sigma_{\min}(\mathbf{V}) + \epsilon)$
    \State \quad $\tilde{R}_t \gets (R_t^{\text{lower}} + R_t^{\text{upper}}) / 2$
\EndFor

\State $\hat{R}_t \gets \text{Softplus}(\mathbf{W}_R \tilde{R}_t + b_R)$ for all $t$

\State \Return $\{\hat{R}_1, \hat{R}_2, \ldots, \hat{R}_T\}$
\end{algorithmic}
\end{algorithm}


\subsection{Epidemic Pre-training and Inference}
\label{sec: train_infer}
\textbf{\underline{Pre-training.}} 
We proposed a dual stage pre-training strategy, where the first stage is to inject epidemic inductive bias into the model via large scale simulation data and the second stage is to adapt to real-world data. 
During first stage of pre-training, for each sample, EpiRecipe provides ground truth for (1) the next token of all compartments and the $R_t$, and (2) a binary compartment mask $\mathbf{M}$ to block the irrelevant compartments. Therefore, the loss function is formulated as:
\begin{equation}
    \mathcal{L} = \sum_i^{|C|} \alpha_i M_i \cdot \text{MSE}(\mathbf{Y}_i, \mathbf{\hat{Y}}_i) + \beta \mathcal{L}_{R_t},
\end{equation}
where $\mathbf{\alpha}$ and $\beta$ are weights to control the alignment with the ground truth compartments and the underlying effective reproduction number.
In the second stage, we collect historical infections from different diseases and conduct the same pretraining strategy while only activating S and I compartments and without applying $\mathcal{L}_{R_t}$.

\noindent\textbf{\underline{Stochastic Structural Inference.}}
While pre-training provides labels for each compartment and the compartment mask, \textit{structural heterogeneity} and \textit{hidden population states} remain challenges for downstream real-world samples. Therefore, during inference on downstream datasets, CAPE provides predictions with uncertainty by predicting $K$ times while randomly masking each compartment with a probability of 50\% (Basic compartments like S and I will never be masked). This process implicitly samples $K$ distinct epidemic model structures, producing a distribution of plausible forecasts. The final forecast is derived by aggregating these realizations (e.g., pooling for point prediction and quantiles for uncertainty quantification), which effectively mimics an ensemble of diverse compartmental models~\cite{reich2019accuracy}.

\begin{table*}[t]
\centering
\caption{Zero-shot forecasting performance (MSE / MAE) across 17 diseases. \textbf{Bold}: best, \underline{underline}: second best. First Count: number of diseases where CAPE beats the best baseline (for CAPE) or ranks \#1 among all models (for baselines).}
\vskip -1.4em
\label{tab:main_results}
\resizebox{\textwidth}{!}{%
\begin{tabular}{l|cc|cc|cccccccccccccccccc}
\toprule
 & \multicolumn{4}{c|}{\textbf{CAPE (Ours)}} & \multicolumn{18}{c}{\textbf{Baselines}} \\
\textbf{Disease} & \multicolumn{2}{c}{CAPE} & \multicolumn{2}{c}{CAPE$_{\text{2nd}}$} & \multicolumn{2}{c}{ARIMA} & \multicolumn{2}{c}{SIR} & \multicolumn{2}{c}{GRU} & \multicolumn{2}{c}{EINN} & \multicolumn{2}{c}{EpiDeep} & \multicolumn{2}{c}{DLinear} & \multicolumn{2}{c}{PatchTST} & \multicolumn{2}{c}{N-BEATS} & \multicolumn{2}{c}{PEM} \\
 & MSE & MAE & MSE & MAE & MSE & MAE & MSE & MAE & MSE & MAE & MSE & MAE & MSE & MAE & MSE & MAE & MSE & MAE & MSE & MAE & MSE & MAE \\
\midrule
Poliomyelitis & \textbf{0.235}{\scriptsize$\pm$0.002} & \textbf{0.226}{\scriptsize$\pm$0.001} & \underline{0.244}{\scriptsize$\pm$0.000} & \underline{0.231}{\scriptsize$\pm$0.001} & 0.326 & 0.261 & 0.833 & 0.535 & 0.435 & 0.251 & 0.672 & 0.315 & 0.860 & 0.417 & 0.365 & 0.291 & 0.910 & 0.380 & 0.417 & 0.302 & 0.820 & 0.365 \\
Diphtheria & 0.168{\scriptsize$\pm$0.000} & \underline{0.196}{\scriptsize$\pm$0.000} & \underline{0.163}{\scriptsize$\pm$0.000} & \textbf{0.192}{\scriptsize$\pm$0.000} & \textbf{0.157} & 0.211 & 0.437 & 0.478 & 1.008 & 0.551 & 1.859 & 0.735 & 1.322 & 0.636 & 1.200 & 0.546 & 2.591 & 0.790 & 2.720 & 0.874 & 1.115 & 0.549 \\
Gonorrhea & 0.060{\scriptsize$\pm$0.001} & \underline{0.138}{\scriptsize$\pm$0.001} & \underline{0.059}{\scriptsize$\pm$0.002} & 0.139{\scriptsize$\pm$0.004} & \textbf{0.052} & \textbf{0.128} & 0.652 & 0.758 & 0.116 & 0.241 & 1.760 & 0.788 & 0.877 & 0.784 & 0.157 & 0.304 & 1.333 & 0.752 & 1.077 & 0.659 & 0.333 & 0.390 \\
Hepatitis A & \underline{0.207}{\scriptsize$\pm$0.006} & \underline{0.318}{\scriptsize$\pm$0.005} & \textbf{0.190}{\scriptsize$\pm$0.008} & \textbf{0.294}{\scriptsize$\pm$0.002} & 0.228 & 0.324 & 0.740 & 0.702 & 0.253 & 0.350 & 0.489 & 0.535 & 0.679 & 0.693 & 0.267 & 0.382 & 0.683 & 0.667 & 0.514 & 0.560 & 0.863 & 0.741 \\
Hepatitis B & \textbf{0.058}{\scriptsize$\pm$0.001} & 0.146{\scriptsize$\pm$0.002} & \underline{0.058}{\scriptsize$\pm$0.004} & \underline{0.145}{\scriptsize$\pm$0.004} & 0.061 & \textbf{0.143} & 0.254 & 0.409 & 0.083 & 0.208 & 0.087 & 0.213 & 0.269 & 0.442 & 0.094 & 0.238 & 0.156 & 0.286 & 0.238 & 0.303 & 0.211 & 0.359 \\
Influenza & 0.450{\scriptsize$\pm$0.008} & 0.283{\scriptsize$\pm$0.018} & 0.435{\scriptsize$\pm$0.001} & \textbf{0.191}{\scriptsize$\pm$0.001} & 1.725 & 0.307 & 0.399 & 0.313 & \underline{0.368} & 0.197 & 0.401 & 0.230 & \textbf{0.261} & 0.239 & 0.465 & 0.252 & 0.666 & 0.405 & 0.446 & \underline{0.191} & 0.526 & 0.261 \\
Meningitis & 0.183{\scriptsize$\pm$0.024} & 0.237{\scriptsize$\pm$0.009} & 0.129{\scriptsize$\pm$0.000} & \underline{0.204}{\scriptsize$\pm$0.001} & 0.150 & 0.233 & 0.487 & 0.549 & \underline{0.129} & 0.218 & 0.153 & 0.234 & 0.239 & 0.363 & 0.211 & 0.275 & 0.309 & 0.394 & \textbf{0.103} & \textbf{0.203} & 0.167 & 0.252 \\
Mumps & \underline{0.019}{\scriptsize$\pm$0.000} & \underline{0.062}{\scriptsize$\pm$0.001} & \textbf{0.019}{\scriptsize$\pm$0.001} & \textbf{0.060}{\scriptsize$\pm$0.002} & 0.025 & 0.068 & 0.249 & 0.475 & 0.046 & 0.154 & 0.024 & 0.102 & 0.156 & 0.373 & 0.058 & 0.208 & 0.054 & 0.181 & 0.046 & 0.160 & 0.075 & 0.217 \\
Pertussis & \textbf{0.080}{\scriptsize$\pm$0.000} & \textbf{0.120}{\scriptsize$\pm$0.000} & \underline{0.085}{\scriptsize$\pm$0.000} & \underline{0.123}{\scriptsize$\pm$0.001} & 0.109 & 0.142 & 0.226 & 0.376 & 1.258 & 0.590 & 2.030 & 0.760 & 1.514 & 0.681 & 0.756 & 0.434 & 1.775 & 0.708 & 2.149 & 0.753 & 1.716 & 0.701 \\
Pneumonia & \textbf{0.043}{\scriptsize$\pm$0.000} & \textbf{0.118}{\scriptsize$\pm$0.000} & \underline{0.043}{\scriptsize$\pm$0.000} & \underline{0.120}{\scriptsize$\pm$0.000} & 0.057 & 0.141 & 0.201 & 0.392 & 0.097 & 0.210 & 0.081 & 0.167 & 0.128 & 0.282 & 0.130 & 0.294 & 0.130 & 0.230 & 0.091 & 0.226 & 0.115 & 0.202 \\
Rubella & \underline{0.025}{\scriptsize$\pm$0.001} & \underline{0.080}{\scriptsize$\pm$0.001} & \textbf{0.023}{\scriptsize$\pm$0.000} & \textbf{0.074}{\scriptsize$\pm$0.001} & 0.029 & 0.083 & 0.239 & 0.469 & 0.278 & 0.331 & 0.090 & 0.240 & 0.600 & 0.713 & 0.131 & 0.320 & 2.224 & 1.161 & 1.679 & 0.776 & 1.325 & 0.875 \\
Scarlet Fever & \underline{0.092}{\scriptsize$\pm$0.000} & 0.188{\scriptsize$\pm$0.000} & \textbf{0.088}{\scriptsize$\pm$0.000} & \underline{0.184}{\scriptsize$\pm$0.000} & 0.092 & \textbf{0.178} & 0.273 & 0.425 & 0.263 & 0.330 & 1.154 & 0.659 & 1.707 & 0.836 & 0.339 & 0.378 & 0.750 & 0.530 & 2.159 & 0.948 & 1.325 & 0.735 \\
Smallpox & \underline{0.146}{\scriptsize$\pm$0.000} & \underline{0.194}{\scriptsize$\pm$0.000} & \textbf{0.142}{\scriptsize$\pm$0.001} & \textbf{0.192}{\scriptsize$\pm$0.000} & 0.170 & 0.213 & 0.382 & 0.424 & 0.605 & 0.396 & 1.451 & 0.613 & 0.915 & 0.488 & 0.694 & 0.405 & 0.976 & 0.522 & 0.733 & 0.418 & 0.780 & 0.439 \\
Tuberculosis & 0.207{\scriptsize$\pm$0.003} & \underline{0.294}{\scriptsize$\pm$0.002} & \underline{0.192}{\scriptsize$\pm$0.002} & 0.307{\scriptsize$\pm$0.003} & \textbf{0.173} & \textbf{0.268} & 0.328 & 0.420 & 0.858 & 0.622 & 1.816 & 0.876 & 1.264 & 0.753 & 0.329 & 0.395 & 1.328 & 0.735 & 1.855 & 0.865 & 1.549 & 0.830 \\
Typhoid Fever & 0.485{\scriptsize$\pm$0.000} & 0.320{\scriptsize$\pm$0.000} & \underline{0.473}{\scriptsize$\pm$0.001} & \underline{0.318}{\scriptsize$\pm$0.001} & \textbf{0.330} & \textbf{0.308} & 0.705 & 0.575 & 0.635 & 0.411 & 1.491 & 0.622 & 1.268 & 0.578 & 0.896 & 0.475 & 0.810 & 0.479 & 10.26 & 1.299 & 1.716 & 0.709 \\
Varicella & \underline{0.076}{\scriptsize$\pm$0.000} & \textbf{0.144}{\scriptsize$\pm$0.000} & \textbf{0.076}{\scriptsize$\pm$0.000} & \underline{0.145}{\scriptsize$\pm$0.001} & 0.086 & 0.154 & 0.206 & 0.403 & 0.322 & 0.386 & 0.476 & 0.423 & 0.567 & 0.571 & 0.491 & 0.512 & 0.591 & 0.535 & 0.642 & 0.582 & 0.611 & 0.603 \\
Measles & \underline{0.207}{\scriptsize$\pm$0.001} & \textbf{0.206}{\scriptsize$\pm$0.000} & \textbf{0.206}{\scriptsize$\pm$0.001} & 0.209{\scriptsize$\pm$0.001} & 0.230 & \underline{0.208} & 0.652 & 0.515 & 1.127 & 0.470 & 1.530 & 0.581 & 1.358 & 0.547 & 1.072 & 0.548 & 1.277 & 0.524 & 1.635 & 0.612 & 1.224 & 0.487 \\
\midrule
\textbf{Average} & \underline{0.161} & \underline{0.192} & \textbf{0.154} & \textbf{0.184} & 0.235 & 0.198 & 0.427 & 0.483 & 0.464 & 0.348 & 0.915 & 0.476 & 0.823 & 0.553 & 0.450 & 0.368 & 0.974 & 0.546 & 1.575 & 0.572 & 0.851 & 0.513 \\
\textbf{1st Count} & 11 & 10 & 11 & 10 & 4 & 5 & 0 & 0 & 0 & 0 & 0 & 0 & 1 & 0 & 0 & 0 & 0 & 0 & 1 & 1 & 0 & 0 \\
\bottomrule
\end{tabular}
}%
\vskip -1em
\end{table*}

\begin{table*}[ht]
\centering
\caption{Ablation Study Results (MSE $\pm$ Std). We conduct evaluations on the full data without train/val/test splits. Subscripts show relative improvement. Note that both pre-training methods are compared against the w/o pre-training baseline.}
\vskip -1.4em
\label{tab:ablation}
\resizebox{0.9\textwidth}{!}{%
\begin{tabular}{@{}lccccccc@{}}
\toprule
\textbf{Method} & \textbf{Influenza} & \textbf{Diphtheria} & \textbf{Gonorrhea} & \textbf{Poliomyelitis} & \textbf{Tuberculosis} & \textbf{Rubella} & \textbf{Avg} \\
\midrule
w/o Pre-training & 0.805{\tiny$\pm$0.004} & 2.048{\tiny$\pm$0.003} & 4.018{\tiny$\pm$0.002} & 1.747{\tiny$\pm$0.003} & 2.171{\tiny$\pm$0.003} & 1.135{\tiny$\pm$0.006} & 1.987 \\
+ Pre-training (Real-world) & 0.835{\tiny$\pm$0.032}$_{\textcolor{red}{\scriptsize\uparrow3.8\%}}$ & 1.674{\tiny$\pm$0.032}$_{\textcolor{teal}{\scriptsize\downarrow18.3\%}}$ & 0.491{\tiny$\pm$0.032}$_{\textcolor{teal}{\scriptsize\downarrow87.8\%}}$ & 1.685{\tiny$\pm$0.032}$_{\textcolor{teal}{\scriptsize\downarrow3.6\%}}$ & 1.619{\tiny$\pm$0.032}$_{\textcolor{teal}{\scriptsize\downarrow25.4\%}}$ & 0.116{\tiny$\pm$0.032}$_{\textcolor{teal}{\scriptsize\downarrow89.8\%}}$ & 1.070$_{\textcolor{teal}{\scriptsize\downarrow46.2\%}}$ \\
\phantom{+} Pre-training (Naive Synth) & 0.803{\tiny$\pm$0.039}$_{\textcolor{teal}{\scriptsize\downarrow0.3\%}}$ & 1.481{\tiny$\pm$0.043}$_{\textcolor{teal}{\scriptsize\downarrow27.7\%}}$ & 0.768{\tiny$\pm$0.039}$_{\textcolor{teal}{\scriptsize\downarrow80.9\%}}$ & 1.572{\tiny$\pm$0.042}$_{\textcolor{teal}{\scriptsize\downarrow10.0\%}}$ & 1.379{\tiny$\pm$0.040}$_{\textcolor{teal}{\scriptsize\downarrow36.5\%}}$ & 0.277{\tiny$\pm$0.041}$_{\textcolor{teal}{\scriptsize\downarrow75.6\%}}$ & 1.047$_{\textcolor{teal}{\scriptsize\downarrow47.3\%}}$ \\
+ Comp. Decoding & 0.678{\tiny$\pm$0.100}$_{\textcolor{teal}{\scriptsize\downarrow15.6\%}}$ & 0.549{\tiny$\pm$0.066}$_{\textcolor{teal}{\scriptsize\downarrow62.9\%}}$ & 0.059{\tiny$\pm$0.020}$_{\textcolor{teal}{\scriptsize\downarrow92.3\%}}$ & 0.448{\tiny$\pm$0.097}$_{\textcolor{teal}{\scriptsize\downarrow71.5\%}}$ & 0.603{\tiny$\pm$0.086}$_{\textcolor{teal}{\scriptsize\downarrow56.3\%}}$ & 0.064{\tiny$\pm$0.048}$_{\textcolor{teal}{\scriptsize\downarrow76.8\%}}$ & 0.400$_{\textcolor{teal}{\scriptsize\downarrow61.8\%}}$ \\
+ Multi-groups Attn. & \textbf{0.652}{\tiny$\pm$0.112}$_{\textcolor{teal}{\scriptsize\downarrow3.7\%}}$ & \textbf{0.547}{\tiny$\pm$0.093}$_{\textcolor{teal}{\scriptsize\downarrow0.5\%}}$ & \textbf{0.046}{\tiny$\pm$0.025}$_{\textcolor{teal}{\scriptsize\downarrow22.4\%}}$ & \textbf{0.463}{\tiny$\pm$0.090}$_{\textcolor{red}{\scriptsize\uparrow3.2\%}}$ & \textbf{0.450}{\tiny$\pm$0.092}$_{\textcolor{teal}{\scriptsize\downarrow25.3\%}}$ & \textbf{0.068}{\tiny$\pm$0.040}$_{\textcolor{red}{\scriptsize\uparrow5.9\%}}$ & \textbf{0.371}$_{\textcolor{teal}{\scriptsize\downarrow7.3\%}}$ \\
\bottomrule
\end{tabular}%
}
\vskip -1em
\end{table*}

\begin{figure}[ht]
    \centering
    \includegraphics[width=0.72\linewidth]{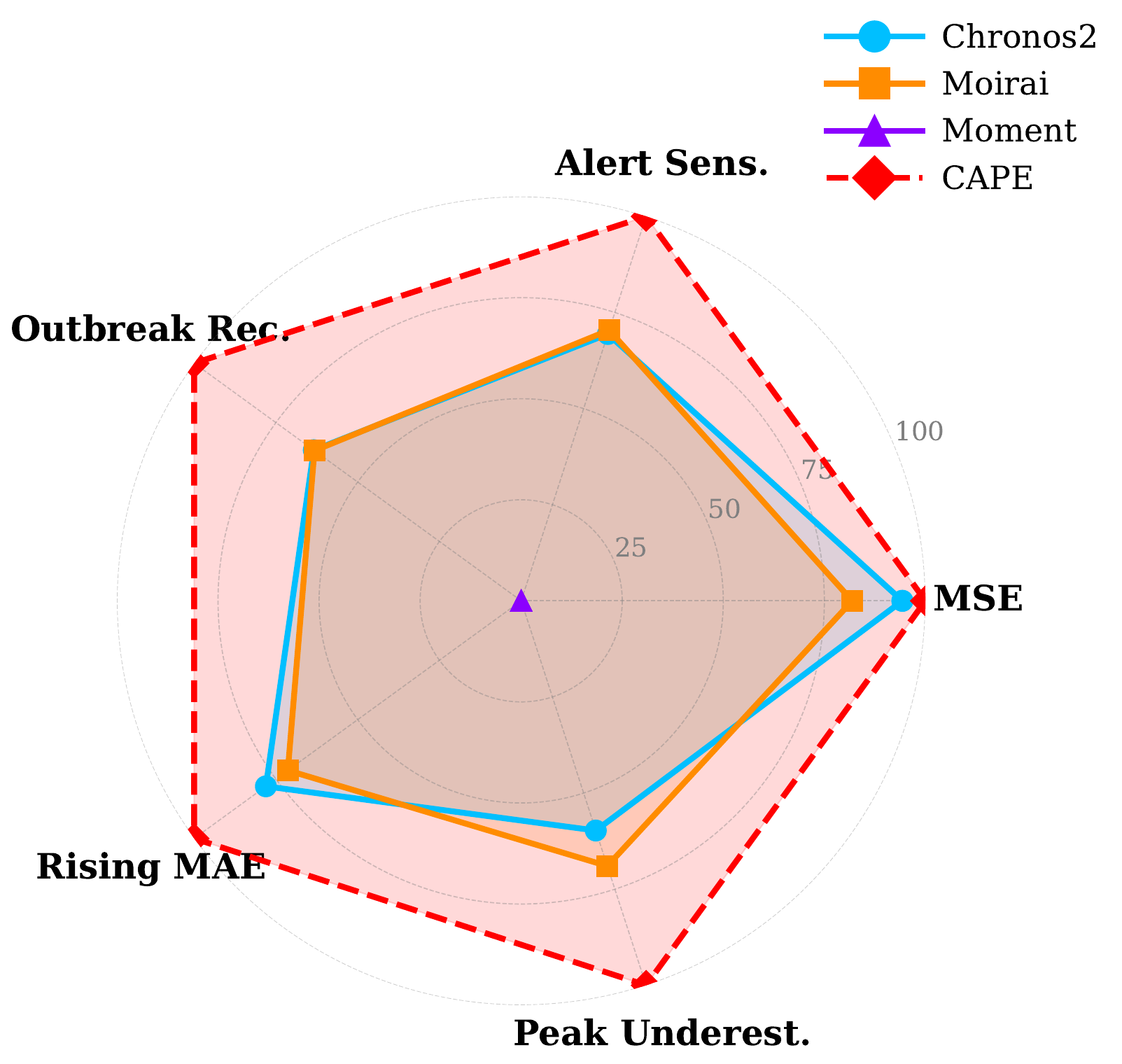}
    \vskip -1.5em
    \caption{Comparison with foundation models across all datasets. Per-disease comparison is shown in Appendix~\ref{app: radar}.}
    \label{fig: fm_comparison}
    \vskip -1.3em
\end{figure}

\vspace{-0.2em}
\section{Experiment}

\subsection{Setup}

\noindent\textbf{Datasets.}
\uline{\textit{Pre-train Corpus:}}
To support pre-training without loading the large-scale pre-compiled datasets. Our pipeline applies a streaming data style by simulating a batch of samples on-the-fly, with padding to match the varying length of time series. We formulate 4 time steps as a token (one month) and pre-train the model with approximately 240M tokens. Then, the second stage pre-training is conducted on 7 distinct real-world diseases with the same token size.
\uline{\textit{Downstream Evaluation Datasets:}}
We collect 17 diverse diseases from Project Tycho~\cite{van2018project} with outbreaks across the US states. We adopt an online setting~\cite{rodriguez2023einns} by using the first 30\% for the base training, and the remaining data for evaluation. More details of the datasets can be found in Appendix~\ref{Appendix: dataset}.

\noindent\textbf{Baselines.}
We adopt baseline models from the comprehensive \textit{EpiLearn} toolkit~\citep{liu2024epilearn}, comparing our model against two categories: \textit{non-pretrained} and \textit{pre-trained} models. \textit{Non-pretrained} baselines including RNN-based epidemic forecasters like EpiDeep and EINNs~\citep{adhikari2019epideep, rodriguez2023einns}, the MLP-based model DLinear~\citep{zeng2023transformers}, and transformer-based architectures like PatchTST~\citep{nie2022time, kamarthi2023pems}. \textit{Pre-trained} baselines include state-of-the-art models such as PEM~\citep{kamarthi2023pems} and the time series foundation model MOMENT~\citep{goswami2024moment}, CHRONOS~\citep{ansari2024chronos} and Moirai~\cite{woo2024unified}. We provide further comparisons with statistical models that are fitted per sample, including ARIMA~\citep{panagopoulos2021transfer} and SIR~\cite{cooper2020sir}. We perform hyperparameter tuning on the hidden sizes, layers, learning rate, weight decay etc, for all baseline models. (See Appendix~\ref{app: implementation} for more details)

\noindent\textbf{Research Questions.} 
In the following experiments, we propose and answer the following questions: 
\textbf{\textit{Q1:}} How does CAPE compare against state-of-the-art non-pretrained and pretrained baselines?
\textbf{\textit{Q2:}} What is the contribution of each component in CAPE's architecture?
\textbf{\textit{Q3:}} What is CAPE learning from the vast pre-training corpus?
\textbf{\textit{Q4:}} Do epidemic forecasting models exhibit scaling laws similar to those observed in language models?
\textbf{\textit{Q5:}} What's the impact of pre-training on forecasting downstream real-world diseases? Specifically: (1) How does the diversity of pre-training data affect downstream performance? (2) What is the relationship between pre-train compute and downstream improvement? (3) Can CAPE learn meaningful disease representations? (4) Can CAPE produce uncertainty quantification?

\subsection{Comparison with Baselines}
To answer \textbf{Q1}, we benchmark CAPE against both full-shot and zero-shot baselines across 17 diseases, demonstrating state-of-the-art performance in both settings.

\noindent\textbf{Comparison with Full-shot Models.}
As shown in Table~\ref{tab:main_results}, both CAPE variants, pre-trained solely on synthetic data (CAPE$_{\text{ZERO}}$) and with additional second-stage pre-training on real-world surveillance data (CAPE), outperform the best baseline on 11/17 diseases in MSE and 10/17 in MAE, without finetuning on the training set. CAPE achieves the best overall performance (MSE: 0.154, MAE: 0.184), substantially outperforming the strongest baseline ARIMA (MSE: 0.235, MAE: 0.198). All deep learning baselines struggle in this data-scarce regime, with several exhibiting catastrophic failures. CAPE's advantages are most pronounced on vaccine-preventable diseases (e.g., Measles, Mumps, Pertussis, Poliomyelitis), reducing MSE by 20--30\% over the best baseline, suggesting that compartmental pre-training effectively captures nonlinear transmission dynamics. ARIMA retains an edge on trend-dominated diseases (e.g., Typhoid Fever), while Influenza remains challenging due to sharp seasonal spikes.

\noindent\textbf{Comparison with Zero-shot Foundation Models.}
We further compare CAPE against state-of-the-art time series foundation models, e.g., Chronos2~\cite{ansari2024chronos}, Moirai~\cite{woo2024unified}, and Moment~\cite{goswami2024moment}, in the zero-shot setting. As shown in Figure~\ref{fig: fm_comparison}, CAPE achieves the lowest MSE and Rising Phase MAE, demonstrating superior point forecast accuracy. More importantly, CAPE excels on epidemiologically critical metrics with the highest Alert Sensitivity (78.35\% vs.\ 64.21\% for Moirai), Outbreak Recall (86.65\% vs.\ 74.04\% for Moirai), and the lowest Peak Underestimate Rate (55.29\% vs.\ 66.99\% for Moirai), indicating better detection of emerging outbreaks and the power of pre-training with domain-specific synthetic data.

\subsection{Ablation Study}
To answer \textbf{Q2},
we evaluate the contribution of each proposed component through an ablation study over 6 diseases, using the entire time series as the test set. As shown in Table~\ref{tab:ablation}, pre-training with naive synthetic data (only the infectious compartment) achieves comparable performance to pre-training with real-world epidemic data (47.3\% vs.\ 46.2\% average improvement), demonstrating that synthetic pre-training can effectively substitute for scarce real-world outbreak data. Adding compartmental decoding provides the most substantial marginal improvement (61.8\%), validating our hypothesis that explicitly modeling latent compartmental states helps capture hidden transmission dynamics across diverse pathogens. The improvement is particularly pronounced for diseases with complex dynamics such as Gonorrhea (92.3\%) and Rubella (76.8\%). The masked multi-groups attention contributes a further 7.3\% average improvement, with notable gains on Tuberculosis (25.3\%) and Gonorrhea (22.4\%), diseases known to exhibit heterogeneous transmission across demographic groups~\cite{trauer2019importance, kirkcaldy2019epidemiology}. Together, these components reduce the average MSE from 1.987 to 0.371, representing an overall improvement of 81.3\%.

\begin{figure*}[t]
    \centering
    \begin{subfigure}[b]{0.24\textwidth}
        \centering
        \includegraphics[width=\linewidth]{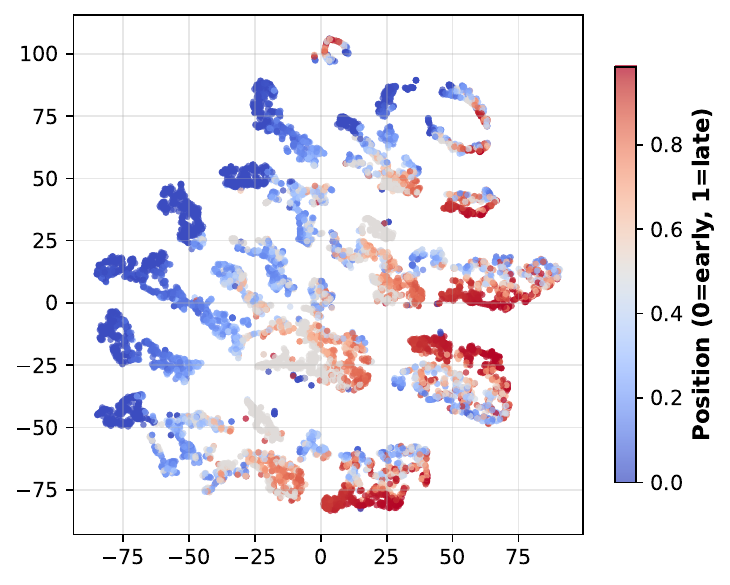}
        \caption{Peak Timing}
        \label{fig:peak_timing}
    \end{subfigure}
    \hfill
    \begin{subfigure}[b]{0.24\textwidth}
        \centering
        \includegraphics[width=\linewidth]{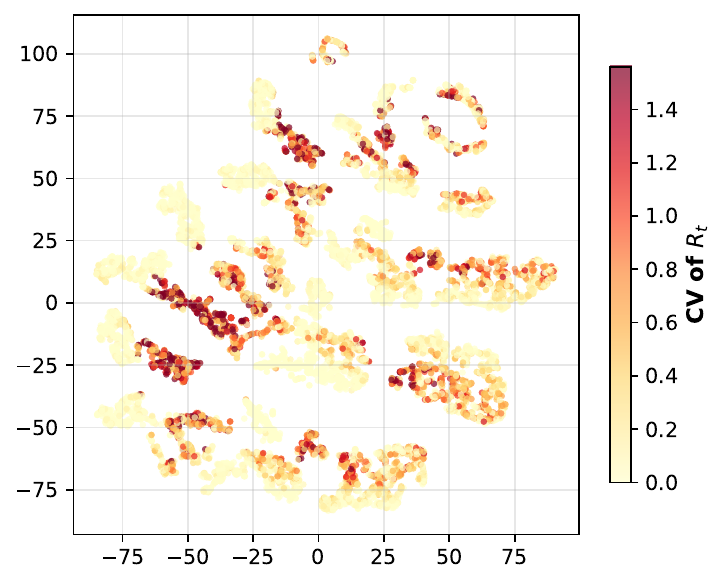}
        \caption{$R_t$ Volatility}
        \label{fig:rt_volatility}
    \end{subfigure}
    \hfill
    \begin{subfigure}[b]{0.24\textwidth}
        \centering
        \includegraphics[width=\linewidth]{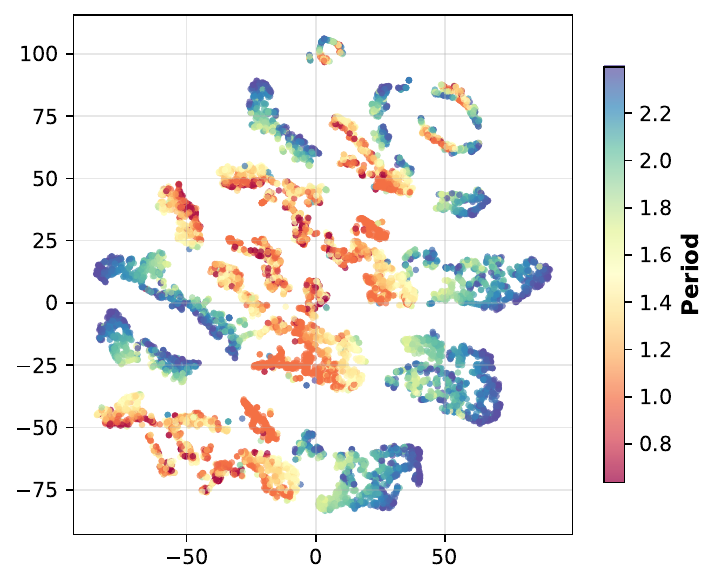}
        \caption{$R_t$ Dominant Period}
        \label{fig:rt_period}
    \end{subfigure}
    \hfill
    \begin{subfigure}[b]{0.24\textwidth}
        \centering
        \includegraphics[width=\linewidth]{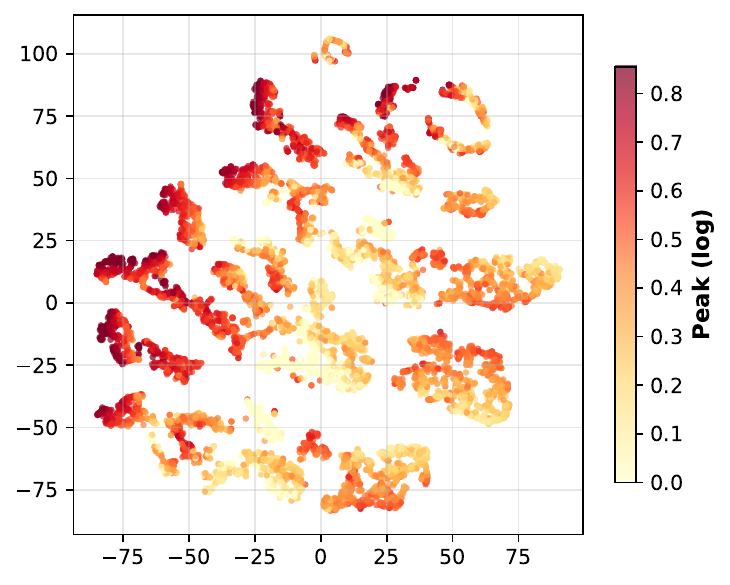}
        \caption{Epidemic Intensity}
        \label{fig:intensity}
    \end{subfigure}
    \vskip -1em
    \caption{t-SNE visualization of CAPE encodings colored by four key epidemiological properties. The visualization reveals that CAPE learns to organize epidemic trajectories into coherent local clusters that correlate strongly with properties such as peak timing, volatility, periodicity, and intensity, capturing both fine-grained temporal dynamics and higher-level structural patterns. More properties are shown in Appendix~\ref{app: rep_prop}.}
    \label{fig: synthetic_embeddings}
    \vskip -1em
\end{figure*}

\subsection{Learning from Simulation Data}
To answer \textbf{Q3}, we investigate the representations learned by CAPE and find that not only EpiRecipe is able to produce infection trajectories with diverse epidemic properties but also CAPE is learn good representation of these properties from the synthetic data.
Specifically, we extract patch embeddings from 10,000 synthetic epidemic trajectories generated by EpiRecipe and visualize them using t-SNE, colored by various eight epidemiological properties (See details in Appendix~\ref{app: rep_prop}), as shown in Figure~\ref{fig: synthetic_embeddings}. The visualization reveals that CAPE learns to organize epidemic trajectories into coherent local clusters that correlate strongly with meaningful epidemiological properties. Notably, the embeddings exhibit smooth gradients for continuous properties (e.g., peak timing, $R_t$ dominant period, etc) and distinct groupings for discrete properties (e.g., wave patterns), suggesting that CAPE captures both fine-grained temporal dynamics and higher-level structural patterns. This organization emerges purely from next-token prediction on diverse compartmental simulations, without explicit supervision on these properties, demonstrating that the pre-training objective naturally induces meaningful epidemic representations.

\subsection{Epidemic Scaling Law}
\begin{figure}[t]
    \centering
    \begin{minipage}{0.49\linewidth}
        \centering
        \includegraphics[width=\linewidth]{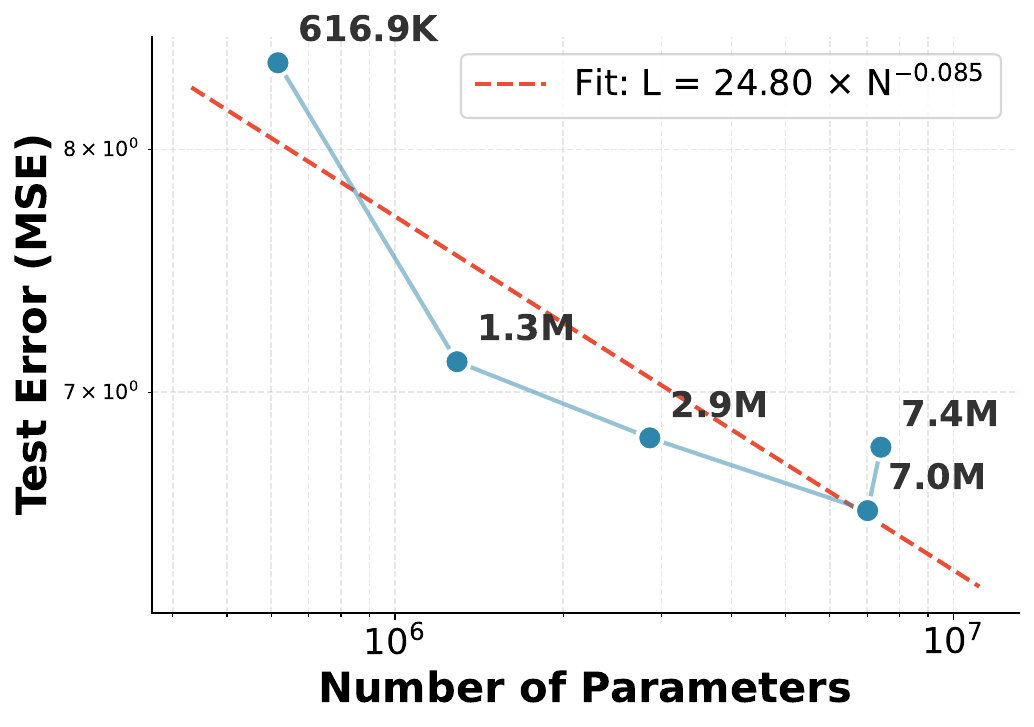}
        \subcaption{MSE vs model size}
        \label{fig:scaling_model_size}
    \end{minipage}\hfill
    \begin{minipage}{0.49\linewidth}
        \centering
        \includegraphics[width=\linewidth]{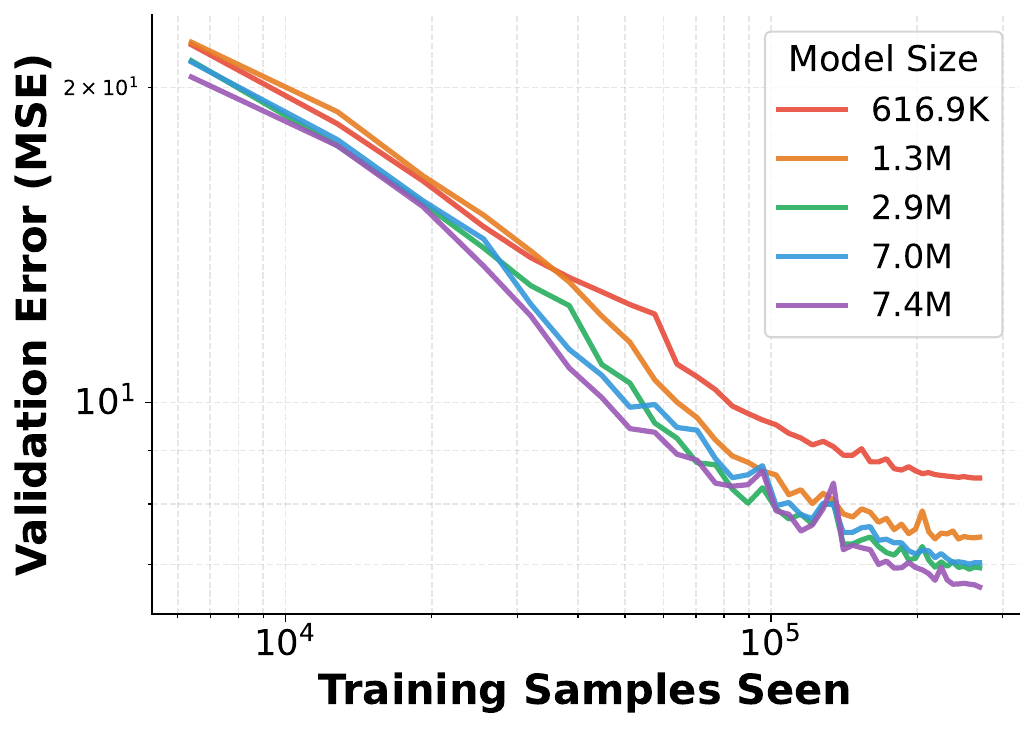}
        \subcaption{MSE vs pre-train compute}
        \label{fig:scaling_pretrain_compute}
    \end{minipage}
    \vskip -1em
    \caption{Scaling under model size and pre-train compute.}
    \label{fig:scaling_overview}
    \vskip -1em
\end{figure}

To answer \textbf{Q4}, 
we investigate whether neural scaling laws~\cite{kaplan2020scaling}, well-established in language and vision domains, also hold for epidemic forecasting. Interestingly, we find that the CAPE models pretrained on EpiRecipe-generated simulations exhibit a certain degree of scaling behavior on both model sizes and the amount of training samples.
\noindent\textbf{(a) Scaling with Model Size.} As shown in Figure~\ref{fig:scaling_model_size}, test error decreases following a power law $L = 24.80 \times N^{-0.085}$ as the number of parameters $N$ increases. The exponent $-0.085$ indicates relatively slow but consistent improvement. This modest scaling exponent suggests that epidemic dynamics, while complex, may have lower intrinsic dimensionality compared to natural language, where exponents typically range from $-0.05$ to $-0.10$~\cite{kaplan2020scaling}.
\noindent\textbf{(b) Scaling with Pre-training Compute.} Figure~\ref{fig:scaling_pretrain_compute} reveals that all model sizes exhibit consistent error reduction as training samples increase, with larger models achieving lower asymptotic error. Interestingly, smaller models (617K--1.3M) show faster initial convergence but plateau earlier, while larger models (7.0M--7.4M) continue improving with more data. This indicates that model capacity and data scale must be balanced: over-parameterized models require sufficient training samples to realize their potential.

\begin{figure}[ht]
    \centering
    \begin{minipage}{0.49\linewidth}
        \centering
        \includegraphics[width=\linewidth]{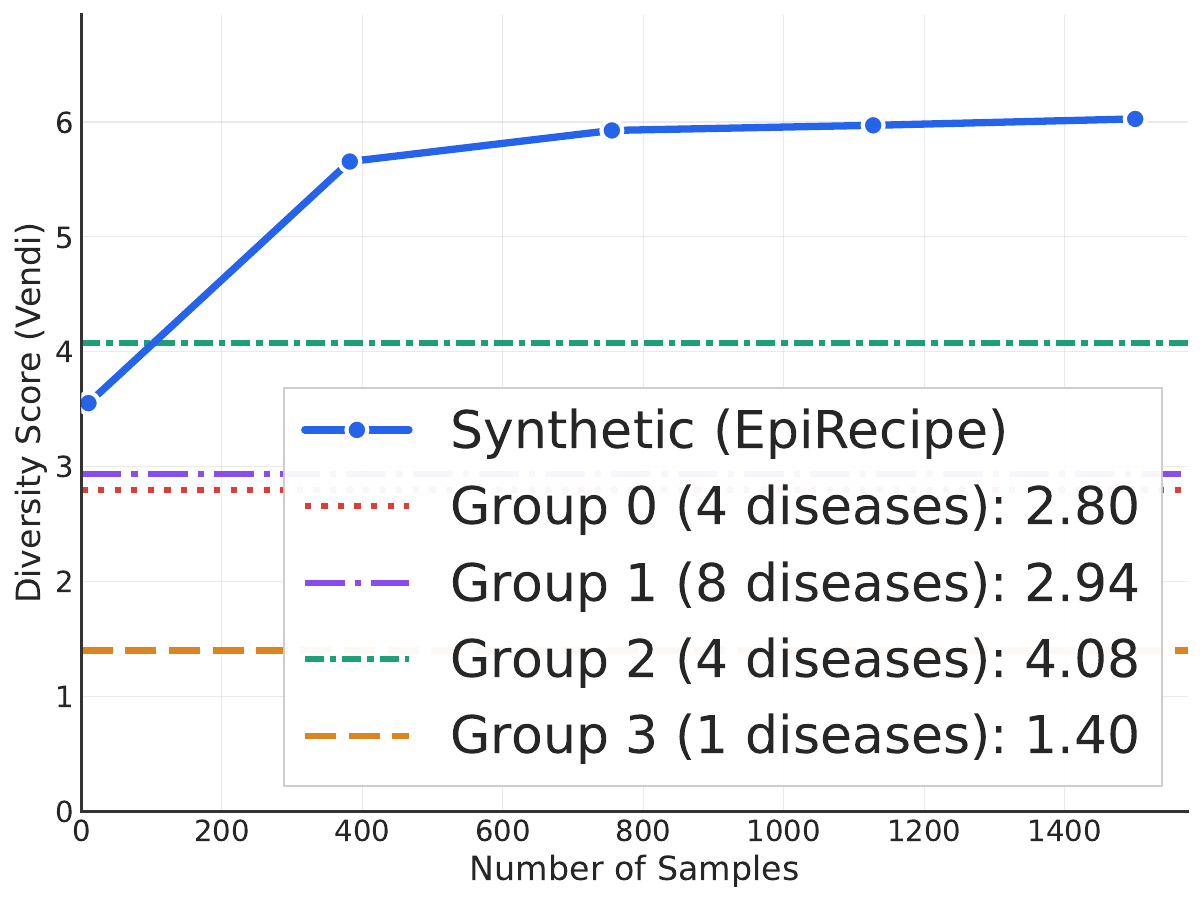}
        \subcaption{Vendi-Scores of each disease group and the synthetic data.}
        \label{fig:vendi}
    \end{minipage}\hfill
    \begin{minipage}{0.49\linewidth}
        \centering
        \includegraphics[width=\linewidth]{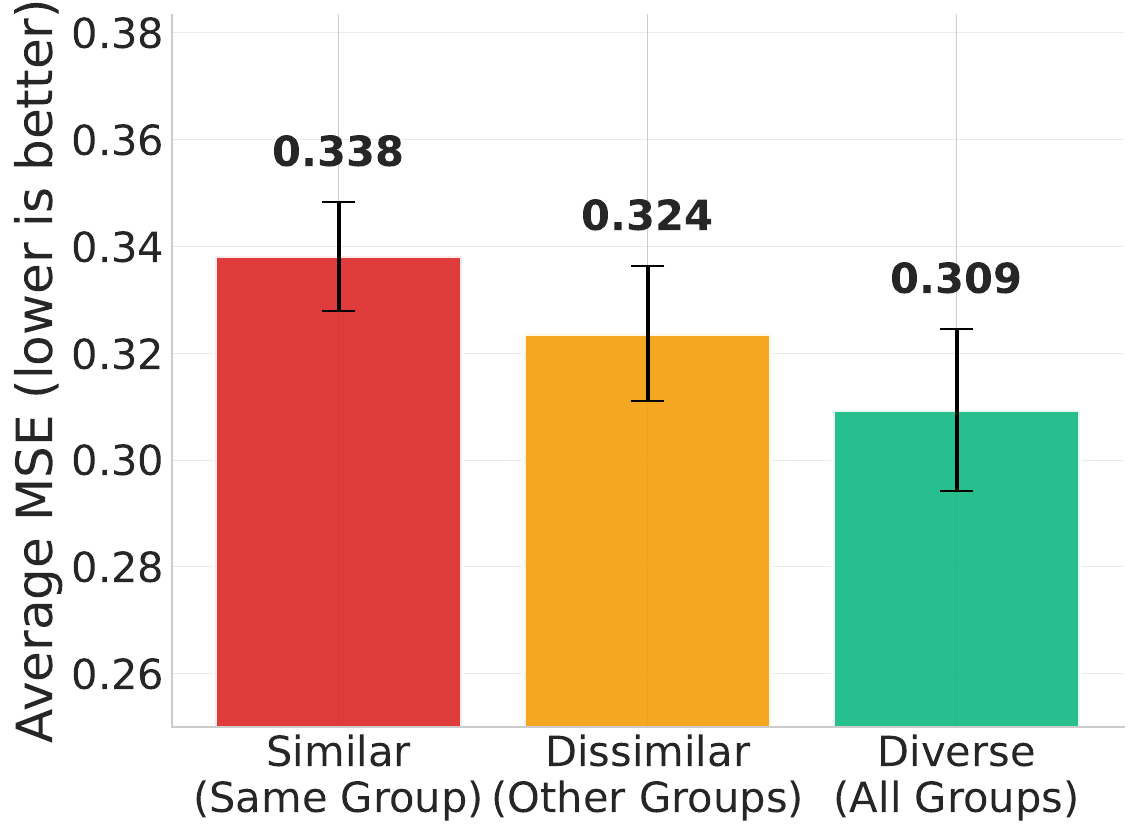}
        \subcaption{Target group MSE with different pre-training corpora.}
        \label{fig:strategies}
    \end{minipage}
    \vskip -1em
    \caption{Diversity scores and transferability experiment.}
    \label{fig:diversity}
    \vskip -1em
\end{figure}

\subsection{From Simulation to Reality: Impact of Pre-training on Downstream Forecasts}
To answer \textbf{Q5}, we explore the effect of epidemic pretraining from both diversity and compute perspectives, and quantified the impact of pre-training over downstream performance, representation quality, and enabling uncertainty estimation.

\noindent\textbf{(a) Impact of Data Diversity on Cross-diseases Transferability.}
We hypothesize that pre-training data diversity, rather than similarity to the target domain, is the key driver of downstream performance. To test this, we cluster 17 diseases from the Tycho dataset into 4 groups based on trajectory features (peak timing, asymmetry, autocorrelation, etc.) using hierarchical clustering, and measure diversity using the Vendi Score~\cite{friedman2022vendi}, which computes the diversity of elements in a dataset based on eigenvalue entropy of the similarity matrix.
To isolate the effect of diversity from data quantity, we conduct a controlled experiment: for each target group, we pre-train on (1) the \textit{same} group (similar), (2) \textit{other} groups excluding the target (dissimilar) group, or (3) \textit{all} groups (most diverse), with the sample count held constant across conditions. As shown in Figure~\ref{fig:strategies}, pre-training on diverse data (All Groups: 0.309 MSE) consistently outperforms pre-training on similar data (Same Group: 0.338 MSE), with an 8.6\% relative improvement. Even pre-training on dissimilar data (Other Groups: 0.324 MSE) outperforms the similar condition, suggesting that exposure to varied dynamics is more valuable than data from the same dynamics. These findings support our design choice of using EpiRecipe's diverse synthetic corpus for pre-training CAPE, which achieves higher diversity as shown in Figure~\ref{fig:vendi}.

\begin{figure}
    \centering
    \includegraphics[width=\linewidth]{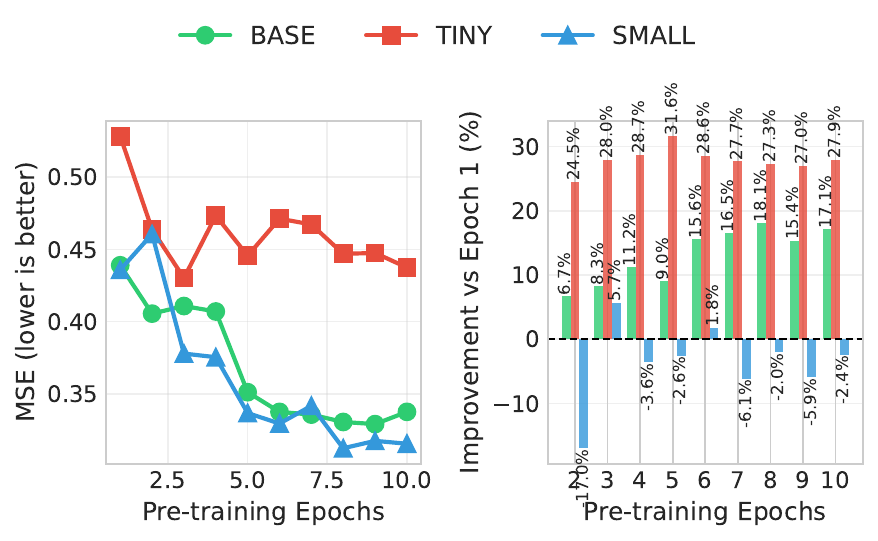}
    \caption{Pre-training compute vs downstream performance. Left: Zero-shot MSE on the Influenza data. Right: Relative improvement over epoch 1 across all diseases.}
    \label{fig:pre-train_compute}
    \vskip -1em
\end{figure}

\noindent\textbf{(b) Pre-training Compute vs Downstream Performance.}
We investigate how pre-training computation affects downstream zero-shot performance by evaluating CAPE checkpoints saved at different epochs on 8 diseases from the Tycho dataset. Figure~\ref{fig:pre-train_compute} (left) shows that MSE decreases monotonically with additional pre-training epochs across all model sizes on the Influenza data. Notably, most improvement occurs in the first 5 epochs, after which gains diminish, consistent with the scaling law findings.
Figure~\ref{fig:pre-train_compute} (right) quantifies the average relative improvement over epoch 1 across all diseases. The BASE model achieves consistent improvements of 15--28\% throughout training, while smaller models like TYNY and SMALL do not show consistent gains or decrease after two epochs, indicating larger models may benefit more from additional compute without overfitting.

\begin{table}[t]
\centering
\caption{Disease Classification (17 disease labels) with linear and non-linear classifiers. (Acc. / F1)}
\vskip -1em
\label{tab:classification}
\small
\resizebox{\columnwidth}{!}{
\begin{tabular}{lccccc}
\toprule
Model & Random & Logistic Reg. & Ridge & KNN & Random Forest \\
\midrule
CAPE & 5.9 / 6 & 22.1 / 9.8 & 23.3 / 11.8 & 55.6 / 50.8 & \textbf{62.7} / \textbf{57.3} \\
PEM & 5.9 / 6 & 56.6 / 51.8 & 48.6 / 39.1 & 64.8 / 61.3 & \textbf{72.9} / \textbf{68.6} \\
\bottomrule
\end{tabular}}
\end{table}

\begin{figure}
    \centering
    \includegraphics[width=\linewidth]{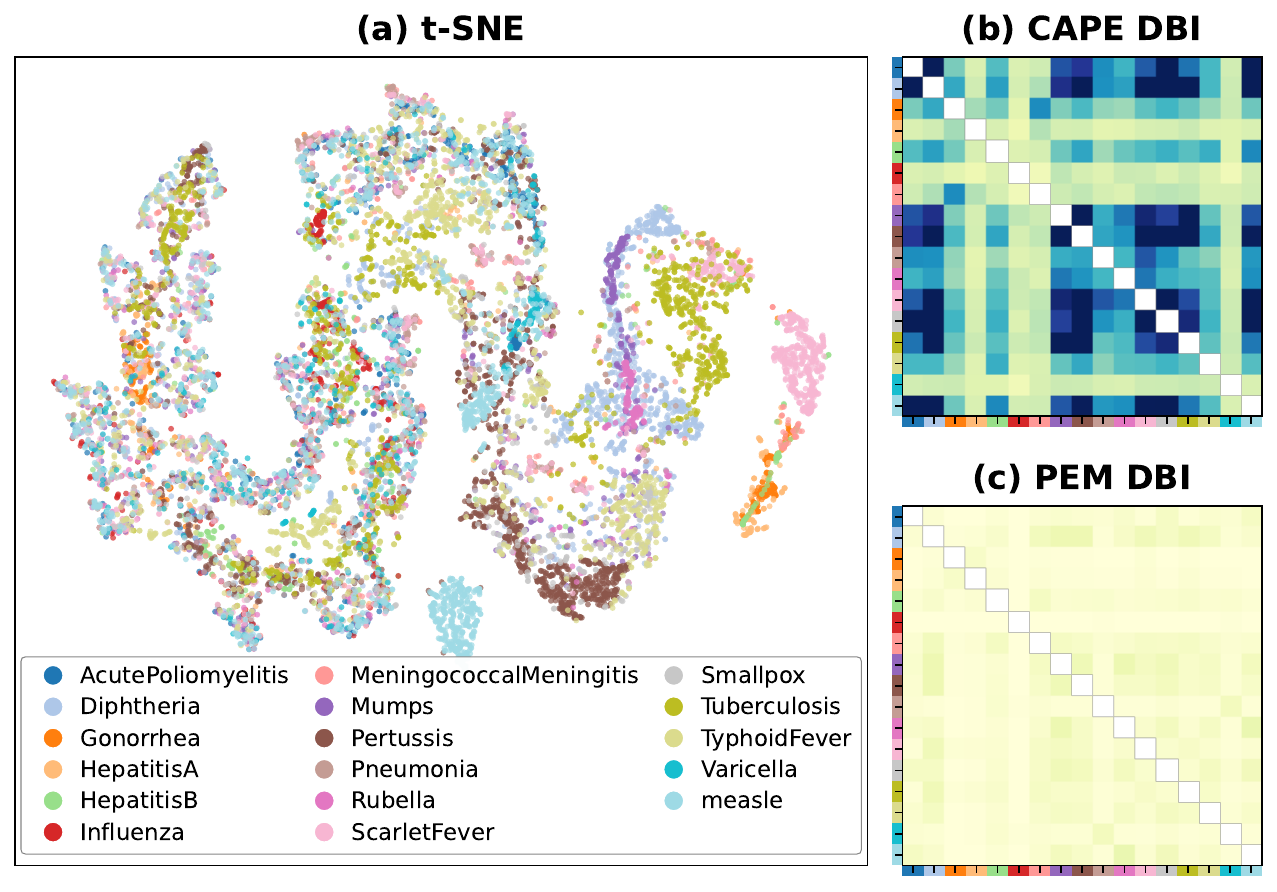}
    \vskip -1em
    \caption{CAPE encodings of downstream diseases and pairwise DBI scores from CAPE and PEM (higher the darker).}
    \label{fig:downstream_encode}
    \vskip -1em
\end{figure}

\noindent\textbf{(c) Learning of Disease Representations.}
Visual analysis via t-SNE (Figure~\ref{fig:downstream_encode}a) reveals overlapping clusters, suggesting CAPE captures shared epidemic dynamics alongside disease-specific features. This balance is supported by pairwise Davies-Bouldin Index (DBI) analysis (Figures~\ref{fig:downstream_encode}b-c), confirming CAPE learns both discriminative and shared patterns. Furthermore, disease classification results (Table~\ref{tab:classification}) show that CAPE representations, unlike PEM, significantly improve non-linear classifier performance, demonstrating the model's ability to encode complex non-linear correlations among diseases.

\begin{figure}
    \centering
    \includegraphics[width=\linewidth]{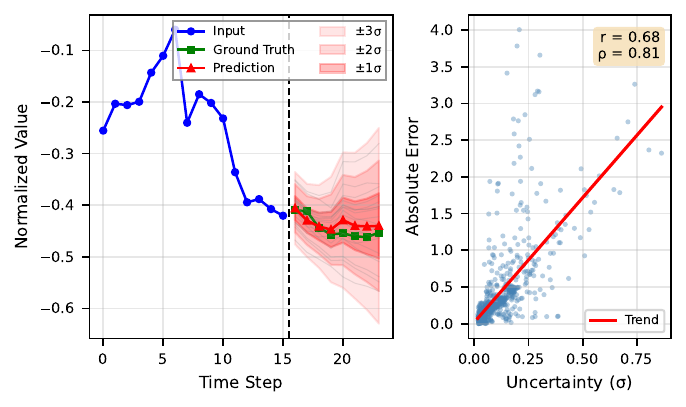}
    \vskip -1em
    \caption{The strong positive correlations (Pearson $r$, Spearman $\rho$) indicate that higher predicted uncertainty corresponds to larger actual errors.}
    \label{fig:uncertainty}
    \vskip -1em
\end{figure}

\noindent\textbf{(d) Uncertainty Modeling via Pre-training.}
CAPE's compartmental masking during pre-training enables uncertainty quantification via stochastic structural inference. By running $K=50$ forward passes with randomly masked compartments, each pass samples a different epidemic model structure. Pooling methods like mean serve as the point forecast, while standard deviation quantifies uncertainty. Figure~\ref{fig:uncertainty} validates calibration quality, as the strong correlations (Pearson $r=0.68$, Spearman $\rho=0.81$) between predicted uncertainty and actual error confirm that the model reliably knows when its predictions are less certain.

\section{Conclusion}
We present CAPE, the first open-source pre-training framework for epidemic forecasting that learns flexible latent population states, termed compartmental prototypes, to address structural heterogeneity and hidden population states in epidemic pre-training. By designing a large-scale pre-training corpus with the proposed EpiRecipe simulation pipeline, CAPE captures generalizable dynamics across diseases. Extensive experiments demonstrate state-of-the-art generalization. In the future, we plan to extend CAPE to incorporate spatial dynamics and textual reasoning abilities for richer generalization, ultimately advancing toward trustworthy and actionable epidemic forecasting.

\bibliographystyle{ACM-Reference-Format}
\section{GenAI Disclosure}
Generative AI tools were employed in this work solely for the purpose of refining the language, correcting grammatical and coding errors.
\bibliography{ref}

\appendix
\clearpage
\section{Appendix}

\subsection{Potential Impact}
\label{app: potential_impact}
This paper advances the interdisciplinary fields of machine learning and epidemiology by enhancing the accuracy of epidemic forecasting in data-limited settings. We outline the potential impacts as follows:

\textbf{Early Insights.}
We provide novel insights into how pre-training and compartment modeling improve epidemic forecasting. Our results demonstrate that pre-training significantly enhances model accuracy, with gains increasing as more pre-training data is incorporated. This finding paves the way for future research to develop foundational models in epidemic analysis using larger datasets. Additionally, we confirm the importance of accounting for both inherent disease dynamics and compartmental factors to achieve robust forecasting performance.

\textbf{Social Impact.}
Epidemic time series data are often sparse due to limited sampling rates, hindering public health organizations' ability to accurately predict infections during novel disease outbreaks. This paper addresses this challenge by showcasing the few-shot and zero-shot forecasting capabilities of pre-trained models. These capabilities can provide powerful tools for early warning and timely intervention, ultimately supporting more effective public health responses and safeguarding communities against emerging infectious diseases.

\subsection{Limitations and Ethical Considerations}
\label{app: limitations}
We use publicly available data without ethical concerns.
While CAPE shows strong performance, it has several limitations. First, the compartment model combinations defined in EpiRecipe can be further expanded. Second, its effectiveness may be constrained by the scale and diversity of available data.

\section{EpiRecipe}
\label{app:epirecipe}

EpiRecipe is a constraint-based pipeline for generating diverse, epidemiologically valid compartmental models. The system features automatic compartment dependency resolution, rule-driven transition construction, hierarchical parameter sampling, and comprehensive model validation. It supports 12 compartment types, 28 transition rules across 14 dynamics categories, and population stratification with configurable mixing patterns. The EpiRecipe pipeline consists of six sequential stages as shown in Figure~\ref{fig:epirecipe_pipeline}. 

\begin{figure}[h]
\centering
\resizebox{0.95\columnwidth}{!}{%
\begin{tikzpicture}[
    node distance=0.5cm and 0.5cm,
    auto,
    block/.style={
        rectangle, 
        draw, 
        fill=blue!10, 
        text width=2.2cm, 
        text centered, 
        rounded corners, 
        minimum height=1.1cm, 
        font=\footnotesize
    },
    arrow/.style={->, thick, >=stealth}
]
    \node[block] (step1) {1. Compartment\\Selection};
    \node[block, right=of step1] (step2) {2. Transition\\Construction};
    
    \node[block, below=of step2] (step3) {3. Parameter\\Sampling};
    \node[block, left=of step3] (step4) {4. Model\\Validation};
    
    \node[block, below=of step4] (step5) {5. Noise\\Injection};
    \node[block, right=of step5] (step6) {6. Simulation};

    \draw[arrow] (step1) -- (step2);
    \draw[arrow] (step2) -- (step3);
    \draw[arrow] (step3) -- (step4);
    \draw[arrow] (step4) -- (step5);
    \draw[arrow] (step5) -- (step6);

\end{tikzpicture}
}
\caption{The EpiRecipe pipeline generates valid compartmental models through constraint satisfaction, hierarchical parameter sampling, and ODEs simulation.}
\label{fig:epirecipe_pipeline}
\end{figure}

\subsection{Compartment Library}

Let $N$ denote the total population size. EpiRecipe supports 12 compartment types, where each compartment $X(t)$ represents the number of individuals in state $X$ at time $t$. Details are provided in Table~\ref{tab:compartments}.

\noindent\textbf{constraint-based selection:} Our pipeline performs constraint-based selection. It first includes the mandatory compartments $\{\textbf{S}, \textbf{I}\} \subseteq \mathcal{C}^*$, then iteratively adds optional compartments while automatically resolving their dependencies, ensuring a valid structure as $\forall c \in \mathcal{C}^*: f_\text{req}(c) \subseteq \mathcal{C}^*$.

\begin{table}[h]
\centering
\caption{Compartment types and structural constraints. Required compartments (S, I) are always included; optional compartments are sampled with automatic dependency resolution. ``Cons.'' indicates population-conserved compartments.}
\label{tab:compartments}
\scriptsize
\setlength{\tabcolsep}{4pt}
\begin{tabular}{@{}clp{4cm}cc@{}}
\toprule
\textbf{Symbol} & \textbf{Type} & \textbf{Description} & \textbf{Req.} & \textbf{Cons.} \\
\midrule
$S$ & Susceptible & Individuals who can contract the disease & $\checkmark$ & $\checkmark$ \\
$I$ & Infectious & Symptomatic individuals who transmit disease & $\checkmark$ & $\checkmark$ \\
$E$ & Exposed & Infected but not yet infectious (latent period) & -- & $\checkmark$ \\
$R$ & Recovered & Individuals with immunity after infection & -- & $\checkmark$ \\
$A$ & Asymptomatic & Infectious individuals without symptoms & -- & $\checkmark$ \\
$H$ & Hospitalized & Severe cases requiring medical care & -- & $\checkmark$ \\
$D$ & Deceased & Cumulative disease-induced deaths & -- & -- \\
$V$ & Vaccinated & Individuals protected by vaccination & -- & $\checkmark$ \\
$Q$ & Quarantined & Isolated individuals (from $I$ or $E$) & -- & $\checkmark$ \\
$W$ & Reservoir & Environmental pathogen concentration & -- & -- \\
$C$ & Chronic & Long-term carriers with prolonged infectivity & -- & $\checkmark$ \\
$P$ & Prophylaxis & Temporarily protected individuals & -- & $\checkmark$ \\
\bottomrule
\end{tabular}
\end{table}

\noindent\textbf{Population Conservation:} Let $\mathcal{L} = \{S, E, I, A, R, H, V, Q, C, P\}$ denote the set of ``living'' compartments. The total living population satisfies $\sum_{X \in \mathcal{L}} X(t) = N - D(t)$, where $D(t)$ is cumulative deaths. The reservoir $W$ represents environmental pathogen load and is excluded from conservation. 

\subsection{Transition Rules}

EpiRecipe implements 28 transition rules organized into 14 dynamics categories. Each rule specifies a flow rate between compartments, as shown in Table~\ref{tab:compartments}.

\begin{table}[h]
\centering
\caption{Complete transition dynamics catalog (28 rules). ``Req.'' indicates compartments that must be present for the rule to apply. All rate parameters are defined in Table~\ref{tab:params_secondary}.}
\label{tab:transitions}
\scriptsize
\setlength{\tabcolsep}{3pt}
\begin{tabular}{@{}p{1.8cm}lp{3.5cm}l@{}}
\toprule
\textbf{Category} & \textbf{Rule Name} & \textbf{Flow Rate Equation} & \textbf{Req.} \\
\midrule
\multirow{7}{*}{\textit{Infection (7)}} 
 & mass\_action & $\beta \cdot S \cdot I / N$ & -- \\
 & with\_asymptomatic & $\beta \cdot S \cdot (I + \epsilon A) / N$ & $A$ \\
 & with\_chronic & $\beta \cdot S \cdot (I + \delta_C C) / N$ & $C$ \\
 & environmental & $\beta_W \cdot S \cdot W / (K + W)$ & $W$ \\
 & combined\_env & $\beta \cdot S \cdot (I + \rho W) / N$ & $W$ \\
 & hospital\_nosocomial & $\beta \cdot S \cdot (I + \theta H) / N$ & $H$ \\
 & saturating & $\beta \cdot S \cdot I / (N(1 + \alpha I))$ & -- \\
\midrule
\multirow{2}{*}{\textit{Latent prog. (2)}} 
 & simple & $\sigma \cdot E \to I$ & $E$ \\
 & branching & $\sigma E \to p_{\text{sym}} I + (1{-}p_{\text{sym}}) A$ & $E,A$ \\
\midrule
\multirow{2}{*}{\textit{Recovery (2)}} 
 & constant\_rate & $\gamma \cdot X$ (where $X \in \{I,A,H\}$) & -- \\
 & resource\_limited & $\gamma \cdot X / (1 + \omega X)$ & -- \\
\midrule
\multirow{2}{*}{\textit{Hospitalization (2)}} 
 & constant\_rate & $\eta \cdot I \to H$ & $H$ \\
 & severity\_dependent & $\eta \cdot I \cdot (1 + \kappa I/N) \to H$ & $H$ \\
\midrule
\multirow{2}{*}{\textit{Death (2)}} 
 & from\_infectious & $\mu \cdot I \to D$ & -- \\
 & from\_hospital & $\mu_H \cdot H \to D$ & $H$ \\
\midrule
\multirow{2}{*}{\textit{Vaccination (2)}} 
 & constant\_campaign & $\nu$ (fixed rate) $\to V$ & $V$ \\
 & proportional & $p_{\text{vax}} \cdot S \to V$ & $V$ \\
\midrule
\multirow{2}{*}{\textit{Quarantine (2)}} 
 & from\_infectious & $\tau_Q \cdot I \to Q$ & $Q$ \\
 & from\_exposed & $\tau_Q \cdot E \to Q$ & $Q,E$ \\
\midrule
\multirow{2}{*}{\textit{Q release (2)}} 
 & to\_recovered & $\rho_Q \cdot Q \to R$ & $Q,R$ \\
 & to\_susceptible & $\rho_Q \cdot Q \to S$ & $Q$ \\
\midrule
\multirow{2}{*}{\textit{Shedding (2)}} 
 & from\_infectious & $\xi \cdot I \to W$ & $W$ \\
 & with\_asymptomatic & $\xi \cdot (I + \delta_A A) \to W$ & $W,A$ \\
\midrule
\textit{Env. decay (1)} & decay & $-\zeta \cdot W$ & $W$ \\
\midrule
\textit{Chronic prog. (1)} & constant & $\chi \cdot I \to C$ & $C$ \\
\midrule
\textit{Chronic clear. (1)} & constant & $\gamma_C \cdot C \to R$ & $C$ \\
\midrule
\textit{Waning imm. (1)} & constant & $\omega \cdot R \to S$ & $R$ \\
\midrule
\textit{Vaccine wane (1)} & constant & $\omega_V \cdot V \to S$ & $V$ \\
\bottomrule
\end{tabular}
\end{table}

\subsection{Hierarchical Parameter Sampling}
\label{sec:params}

Parameters are sampled in three hierarchical tiers to ensure epidemiological consistency:

\begin{enumerate}[itemsep=2pt, topsep=4pt, leftmargin=*]
    \item \textbf{Tier 1 -- Primary Parameters} (Table~\ref{tab:params_primary}): Core epidemiological quantities sampled from informed prior distributions.
    \item \textbf{Tier 2 -- Derived Parameters} (Table~\ref{tab:params_derived}): Computed deterministically from Tier 1 to maintain mathematical consistency.
    \item \textbf{Tier 3 -- Secondary Parameters} (Table~\ref{tab:params_secondary}): Transition-specific rates sampled uniformly within plausible ranges.
\end{enumerate}

\begin{table}[h]
\centering
\caption{Tier 1: Primary parameters sampled from epidemiologically-motivated distributions. Values are clipped to the specified range after sampling.}
\label{tab:params_primary}
\scriptsize
\setlength{\tabcolsep}{4pt}
\begin{tabular}{@{}llll@{}}
\toprule
\textbf{Parameter} & \textbf{Description} & \textbf{Distribution} & \textbf{Range} \\
\midrule
$R_0$ & Basic reproduction number & LogNormal$(\mu{=}2.5, \sigma{=}0.8)$ & $[1.1, 18.0]$ \\
$T_{\text{inf}}$ & Infectious period (days) & Gamma$(k{=}3.0, \theta{=}3.0)$ & $[3, 21]$ \\
$T_{\text{lat}}$ & Latent period (days, if $E$ present) & Gamma$(k{=}2.0, \theta{=}2.5)$ & $[1, 14]$ \\
\bottomrule
\end{tabular}
\end{table}

\begin{table}[h]
\centering
\caption{Tier 2: Derived parameters computed deterministically from Tier~1 to ensure epidemiological consistency (e.g., $\beta = R_0 \cdot \gamma$).}
\label{tab:params_derived}
\scriptsize
\setlength{\tabcolsep}{4pt}
\begin{tabular}{@{}lll@{}}
\toprule
\textbf{Parameter} & \textbf{Formula} & \textbf{Description} \\
\midrule
$\gamma$ & $1 / T_{\text{inf}}$ & Recovery rate (inverse of infectious period) \\
$\sigma$ & $1 / T_{\text{lat}}$ & Latent progression rate (inverse of latent period) \\
$\beta$ & $R_0 \cdot \gamma$ & Transmission rate (ensures correct $R_0$) \\
\bottomrule
\end{tabular}
\end{table}

\begin{table}[h]
\centering
\caption{Tier 3: Secondary parameters (24 total) sampled uniformly within biologically plausible ranges. These govern transition-specific dynamics.}
\label{tab:params_secondary}
\scriptsize
\setlength{\tabcolsep}{3pt}
\begin{tabular}{@{}llp{5.5cm}@{}}
\toprule
\textbf{Param.} & \textbf{Range} & \textbf{Description} \\
\midrule
$\epsilon$ & $[0.3, 0.9]$ & Relative infectiousness of asymptomatic vs symptomatic \\
$\delta_C$ & $[0.1, 0.5]$ & Relative infectiousness of chronic carriers \\
$p_{\text{sym}}$ & $[0.3, 0.8]$ & Fraction of exposed becoming symptomatic \\
$\eta$ & $[0.01, 0.15]$ & Rate of hospitalization from $I$ \\
$\mu$ & $[0.001, 0.03]$ & Disease-induced mortality rate from $I$ \\
$\mu_H$ & $[0.02, 0.15]$ & Mortality rate from hospital $H$ \\
$\gamma_H$ & $[0.05, 0.2]$ & Hospital discharge rate \\
$\gamma_A$ & $[0.1, 0.3]$ & Recovery rate for asymptomatic \\
$\gamma_C$ & $[0.005, 0.05]$ & Clearance rate for chronic carriers \\
$\chi$ & $[0.01, 0.1]$ & Rate of developing chronic carriage \\
$\omega$ & $[0.001, 0.02]$ & Immunity waning rate ($R \to S$) \\
$\omega_V$ & $[0.002, 0.01]$ & Vaccine waning rate ($V \to S$) \\
$p_{\text{vax}}$ & $[0.001, 0.02]$ & Proportional vaccination rate \\
$\nu$ & $[10, 500]$ & Constant vaccination rate (individuals/time) \\
$\tau_Q$ & $[0.01, 0.1]$ & Quarantine rate \\
$\rho_Q$ & $[0.05, 0.15]$ & Quarantine release rate \\
$\xi$ & $[0.001, 0.1]$ & Environmental shedding rate \\
$\delta_A$ & $[0.3, 1.0]$ & Relative shedding from asymptomatic \\
$\zeta$ & $[0.05, 0.5]$ & Environmental decay rate \\
$\beta_W$ & $[0.001, 0.1]$ & Environmental transmission rate \\
$K$ & $[100, 10000]$ & Half-saturation constant for environmental transmission \\
$\rho$ & $[0.1, 1.0]$ & Environmental contribution weight \\
$\theta$ & $[0.1, 0.5]$ & Nosocomial transmission factor \\
$\alpha$ & $[0.0001, 0.01]$ & Behavioral saturation parameter \\
\bottomrule
\end{tabular}
\end{table}

\noindent\textbf{Seasonal Forcing} (applied to 50\% of models): The transmission rate varies periodically as $\beta(t) = \beta_0 (1 + \varepsilon \cos(2\pi t/T + \phi))$, where $\varepsilon \in [0.1, 0.5]$ is the amplitude, $T \in \{26, 52, 104, 156\}$ weeks is the period, and $\phi \sim \text{Uniform}(0, 2\pi)$ is the phase offset.
\subsection{Model Validation}

Every generated model is validated against five invariants: (1) population conservation, (2) no dead-ends (every non-terminal compartment has outflow), (3) infectious pathway exists from $S$ to $\{I, A, C\}$, (4) $R_0 > 0$, and (5) numerical stability (all rates $\leq 10.0$). Models failing any check are rejected and regenerated.

\subsection{Population Stratification}

EpiRecipe supports stratification by age, risk, spatial, or custom groupings. Each group $g$ has: population fraction $f_g$ ($\sum_g f_g = 1$), contact multiplier $c_g$, and susceptibility multiplier $s_g$..

\begin{table}[h]
\centering
\caption{Preset population stratifications with group-specific multipliers and default mixing patterns.}
\label{tab:stratification}
\scriptsize
\setlength{\tabcolsep}{4pt}
\begin{tabular}{@{}llcccl@{}}
\toprule
\textbf{Preset} & \textbf{Group} & $f_g$ & $c_g$ & $s_g$ & \textbf{Default Mixing} \\
\midrule
\multirow{3}{*}{Age (3 groups)} 
 & Children & 0.20 & 1.5 & 0.8 & \multirow{3}{*}{Assortative ($a{=}0.7$)} \\
 & Adults & 0.60 & 1.0 & 1.0 & \\
 & Elderly & 0.20 & 0.6 & 1.5 & \\
\midrule
\multirow{2}{*}{Risk (2 groups)} 
 & Low-risk & 0.85 & 1.0 & 1.0 & \multirow{2}{*}{Homogeneous} \\
 & High-risk & 0.15 & 1.0 & 1.2 & \\
\bottomrule
\end{tabular}
\end{table}

\noindent\textbf{Mixing Matrices:} The matrix $\mathbf{M} \in \mathbb{R}^{n \times n}$ specifies contact patterns, where $M_{ij}$ is the relative contact rate from group $i$ to $j$ (rows sum to 1):
\begin{itemize}[itemsep=1pt, topsep=2pt, leftmargin=*]
    \item \textbf{Homogeneous}: $M_{ij} = 1/n$ (equal contact probability)
    \item \textbf{Assortative}: $M_{ii} = a$, $M_{ij} = (1{-}a)/(n{-}1)$ for $i \neq j$ (default $a{=}0.7$)
    \item \textbf{Hierarchical}: $M_{ij} \propto e^{-|i-j|/\sigma}$ (age-like decay with distance)
\end{itemize}

\subsection{Simulation}

Models are integrated using RK4 with time step $\Delta t = 1.0$, duration 250 steps, and population $N \in [1000, 100000]$. Post-step rescaling enforces population conservation. Gaussian observation noise with $\sigma_{\text{noise}} \sim \text{Uniform}(0, 0.05)$ is injected. Computed observables include incidence, prevalence ($I + A + C$), $R_t = R_0 \cdot S(t)/N$, hospitalizations, and cumulative deaths.

\section{Theoretical Analysis}
\label{Appendix: theory}

\subsection{Estimation of $R_t$}
\label{app: R_t inference}
\paragraph{Epidemiological Interpretation of Attention Patterns.}
The construction of $\mathbf{F}$ and $\mathbf{V}$ matrices from attention scores follows the next-generation matrix framework for computing $R_t$~\cite{diekmann2010construction}. In this framework, $\mathbf{F}$ represents new infection rates (transmission) and $\mathbf{V}$ represents transition rates out of infectious states (removal), with $R_t = \rho(\mathbf{FV}^{-1})$ where $\rho(\cdot)$ denotes the spectral radius.

\textbf{F Matrix (Transmission):} The off-diagonal elements of the I-to-I attention submatrix $A^{(I)}_{ij}$ ($i \neq j$) capture how strongly infectious group $j$ influences group $i$. This directly parallels transmission rates $\beta_{ij}$ in heterogeneous compartmental models, where infections spread \textit{between} different population groups. We zero the diagonal since self-infection is not meaningful in this context.

\textbf{V Matrix (Removal):} The diagonal elements $A^{(I)}_{ii}$ represent self-attention—how much each infectious group attends to itself. In attention mechanisms, higher self-attention indicates that the token's representation is more self-contained and persistent across layers. This is analogous to the \textit{mean infectious period} $1/\gamma$ in epidemiology: a longer infectious period means individuals remain in the infectious state longer before transitioning out (via recovery, hospitalization, or death).

The inverse relationship $\gamma_i = 1/(A^{(I)}_{ii} + \epsilon)$ arises because:
\begin{itemize}
    \item High self-attention $A_{ii}$ $\rightarrow$ state persists longer $\rightarrow$ low removal rate $\gamma_i$
    \item Low self-attention $A_{ii}$ $\rightarrow$ rapid state transition $\rightarrow$ high removal rate $\gamma_i$
\end{itemize}

This mirrors the classical SIR relationship where $R = \beta/\gamma$: the reproduction number increases with transmission rate $\beta$ (off-diagonal attention) and decreases with removal rate $\gamma$ (inverse of diagonal attention). By computing spectral bounds on $\mathbf{FV}^{-1}$, we obtain time-varying $R(t)$ estimates that reflect learned epidemic dynamics without requiring explicit compartmental supervision.

\subsection{Proof of spectral bounds in Algo.~\ref{alg:rt_estimation}}
\label{app: bound proof}
\newcommand{\C}{\mathbb{C}}
\newcommand{\norm}[1]{\left\lVert #1 \right\rVert}
\newcommand{\ip}[2]{\left\langle #1,\,#2 \right\rangle}

\textbf{Setup and Preliminaries.}
Let $V\in \C^{m\times n}$. 
The (operator) $2$-norm of a matrix $M$ is
\[
\norm{M}_2 \;=\; \sup_{x\neq 0}\frac{\norm{Mx}_2}{\norm{x}_2}
\;=\; \max_{\norm{x}_2=1}\norm{Mx}_2.
\]
The \emph{singular values} $\sigma_1(V)\ge \cdots \ge \sigma_r(V)\ge 0$ (with $r=\operatorname{rank}(V)$) are, by definition, the nonnegative square roots of the eigenvalues of $V^\ast V$ (where ${}^\ast$ denotes conjugate transpose), counted with multiplicity and ordered nonincreasingly.

\medskip

For any matrix $M$ and any vector $x$,
\[
\sigma_{\min}(M)\,\|x\|_2 \;\le\; \|Mx\|_2 \;\le\; \sigma_{\max}(M)\,\|x\|_2,
\quad
\|M\|_2=\sigma_{\max}(M).
\]
For square $M$, the singular values $\{\sigma_i(M)\}_{i=1}^n$ and eigenvalues $\{\lambda_i(M)\}_{i=1}^n$ satisfy
\[
\prod_{i=1}^n \sigma_i(M)
= \sqrt{\det(M^*M)} = |\det M|
= \prod_{i=1}^n |\lambda_i(M)|.
\]
In particular, by comparing geometric means to extrema,\
\[
\small
\begin{aligned}
\sigma_{\min}(M) 
\;\le\; \Big(\prod_{i=1}^n \sigma_i(M)\Big)^{1/n}
= \Big(\prod_{i=1}^n |\lambda_i(M)|\Big)^{1/n}
\;\le\; \max_i |\lambda_i(M)| = \rho(M).
\end{aligned}
\tag{P1}
\]
Thus, for any square $M$,
\[
\sigma_{\min}(M) \;\le\; \rho(M) \;\le\; \sigma_{\max}(M)=\|M\|_2.
\tag{P2}
\]

\begin{theorem}[Bounds for $R_t$]
Let $F,V \in \mathbb{C}^{n\times n}$ with $V$ invertible, and define
\[
R_t \;\equiv\; \rho(FV^{-1}),
\]
where $\rho(\cdot)$ denotes the spectral radius, $\|\cdot\|_2$ the operator $2$-norm, and
$\sigma_{\max}(\cdot), \sigma_{\min}(\cdot)$ the maximal and minimal singular values, respectively.  
Then $R_t$ satisfies the bounds
\[
\boxed{\;
\frac{\sigma_{\min}(F)}{\sigma_{\max}(V)}
\;\le\;
\rho(FV^{-1})
\;\le\;
\frac{\sigma_{\max}(F)}{\sigma_{\min}(V)}.
\;}
\]
\end{theorem}

\medskip
\noindent\textbf{Derivation for Lower bound.}
\begin{proof}
For any compatible $A,B$ and any unit vector $x$,
\[
\|ABx\|_2 \;\ge\; \sigma_{\min}(A)\,\|Bx\|_2 \;\ge\; \sigma_{\min}(A)\,\sigma_{\min}(B)\,\|x\|_2,
\]
hence
\[
\sigma_{\min}(AB) \;\ge\; \sigma_{\min}(A)\,\sigma_{\min}(B).
\tag{S1}
\]
Apply (S1) with $A=F$ and $B=V^{-1}$ (with $V$ invertible):
\[
\sigma_{\min}(FV^{-1}) \;\ge\; \sigma_{\min}(F)\,\sigma_{\min}(V^{-1}).
\]
Using the inversion identity for singular values,
\[
\sigma_{\min}(V^{-1})=\frac{1}{\sigma_{\max}(V)},
\]
we obtain
\[
\boxed{\;\sigma_{\min}(FV^{-1}) \;\ge\; \dfrac{\sigma_{\min}(F)}{\sigma_{\max}(V)}.\;}
\tag{L1}
\]

\medskip
By (P1) applied to $M=FV^{-1}$, we get:
\[
\sigma_{\min}(FV^{-1}) \;\le\; \rho(FV^{-1}).
\tag{L2}
\]
Combining (L1) and (L2) yields the rigorous lower bound
\[
\boxed{\;\dfrac{\sigma_{\min}(F)}{\sigma_{\max}(V)} \;\le\; \rho(FV^{-1}).\;}
\tag{LB}
\]

\end{proof}

\medskip
\noindent\textbf{Derivation for Upper bound.}
\begin{proof}
For any square $M$, $\rho(M)\le \|M\|_2$ by (P2). Thus
\[
\rho(FV^{-1}) \;\le\; \|FV^{-1}\|_2 \;\le\; \|F\|_2\,\|V^{-1}\|_2
= \sigma_{\max}(F)\,\frac{1}{\sigma_{\min}(V)}.
\]
Hence the rigorous upper bound is
\[
\boxed{\;\rho(FV^{-1}) \;\le\; \dfrac{\sigma_{\max}(F)}{\sigma_{\min}(V)}.\;}
\tag{UB}
\]

\medskip
\noindent\textbf{Final result.}
Putting (LB) and (UB) together, we obtain
\[
\boxed{\;
\frac{\sigma_{\min}(F)}{\sigma_{\max}(V)}
\;\le\;
\rho(FV^{-1})
\;\le\;
\frac{\sigma_{\max}(F)}{\sigma_{\min}(V)}.
\;}
\]

\end{proof}

\section{Pre-train and Evaluation Datasets}
\label{Appendix: dataset}

\subsection{Pre-training}

In the first stage, CAPE is pretrained on synthetically generated epidemic time series produced by the EpiRecipe pipeline, which programmatically generates diverse compartmental models with realistic dynamics. This approach enables unlimited pretraining data with known ground truth compartmental structures. In the second stage, we use the following 8 diseases for the same style pre-training: Pertussis, Varicella, Tuberculosis, Measles, TyphoidFever, Mumps, Diphtheria, and ScarletFever. Here, we discuss the details of the first stage pret-train setting as following.

\paragraph{Compartmental Model Generation.}
The EpiRecipe pipeline implements constraint-based model generation with 12 possible compartments: Susceptible (S), Infected (I), Exposed (E), Recovered (R), Hospitalized (H), Vaccinated (V), Quarantined (Q), Deceased (D), Protected (P), Environmental Reservoir (W), Asymptomatic (A), and Chronic carriers (C). We randomly select 3--7 compartments, with S and I as mandatory compartments.

\paragraph{Simulation Parameters.}
Each synthetic epidemic is simulated with:
\begin{itemize}
    \item \textbf{Basic reproduction number}: $R_0 \sim \text{LogNormal}(\mu=2.5, \sigma=0.8)$, clipped to $[1.1, 18.0]$
    \item \textbf{Infectious period}: $\gamma^{-1} \sim \text{Gamma}(k=3, \theta=3)$, ranging 3--21 days
    \item \textbf{Latent period} (if E exists): $\sigma^{-1} \sim \text{Gamma}(k=2, \theta=2.5)$, ranging 1--14 days
    \item \textbf{Population size}: Uniformly sampled from $[10^4, 10^6]$
    \item \textbf{Simulation length}: 52--260 weeks (1--5 years)
    \item \textbf{Time resolution}: Weekly aggregation (default), with optional daily resolution
\end{itemize}

\paragraph{Seasonal Forcing.}
50\% of generated models include seasonal transmission modulation:
\begin{equation}
    \beta(t) = \beta_0 \times \left(1 + \epsilon \cos\left(\frac{2\pi t}{T} + \phi\right)\right)
\end{equation}
where $\epsilon \in [0.1, 0.5]$ is the forcing amplitude, $T \in \{26, 52, 104, 156\}$ weeks represents the period (biannual, annual, 2-year, or 3-year cycles), and $\phi \in [0, 2\pi]$ is a random phase offset. This captures the seasonal patterns observed in respiratory diseases like influenza and measles.

\paragraph{Group-Stratified Models.}
To capture heterogeneous contact patterns, 50\% of training samples use group-stratified models with 2--5 population groups. Each group has distinct:
\begin{itemize}
    \item Population fraction (sampled from Dirichlet distribution)
    \item Contact rate multiplier $\in [0.5, 1.5]$
    \item Susceptibility multiplier $\in [0.8, 1.2]$
    \item Infectivity multiplier $\in [0.8, 1.2]$
\end{itemize}
Mixing patterns include homogeneous, assortative (with assortativity $\in [0.5, 0.9]$), hierarchical, and spatial structures.

\paragraph{Gaussian Process Augmentation.}
Following the KernelSynth approach~\cite{ansari2024chronos}, 20\% of pretraining samples are generated using Gaussian Process (GP) priors with epidemic-relevant kernels:
\begin{itemize}
    \item \textbf{Periodic kernels}: Capturing weekly (7 days), annual (52 weeks), biannual (26 weeks), and multi-year cycles (104--208 weeks)
    \item \textbf{RBF kernels}: Modeling outbreak shapes at various scales (length scales 0.02--0.5)
    \item \textbf{Linear kernels}: Capturing endemic trends
    \item \textbf{Composite kernels}: Random combinations of 1--4 base kernels using $+$ or $\times$ operators
\end{itemize}
This augmentation improves the model's ability to learn diverse periodic patterns commonly observed in epidemic data.

\paragraph{Pretraining Data Volume.}
The streaming data generation produces samples on-the-fly during training, providing effectively unlimited data diversity. Each batch generates 64 synthetic epidemics (configurable), and the pre-training generally takes over 80,000 batches.

\subsection{Downstream Evaluation Datasets}
We evaluate CAPE on 17 real-world disease datasets from Project Tycho~\cite{van2018project}, a database of historical disease incidence in the United States. These datasets span diverse disease types with varying epidemiological characteristics.

\paragraph{Data Processing.}
Each disease dataset is processed as follows:
\begin{enumerate}
    \item \textbf{Aggregation}: Weekly case counts are aggregated across all reporting states to create a single national-level time series, avoiding the complexity of spatial modeling.
    \item \textbf{Normalization}: Time series are standardized using z-score normalization applied on the time series prior to tokenization.
\end{enumerate}

\paragraph{Online Evaluation Protocol.}
We adopt an online (rolling) evaluation setting~\cite{rodriguez2023einns} that simulates real-world deployment scenarios where models must forecast future outbreaks using only historically available data. The evaluation proceeds as follows:
\begin{enumerate}
    \item \textbf{Initial Training Window}: The first 30\% of the tokenized time series is used as the initial training set (base training rate = 0.3).
    \item \textbf{Rolling Folds}: The remaining 70\% of data is divided into $K=2$ non-overlapping evaluation folds. Each fold uses an expanding training window that includes all data up to the fold boundary.
    \item \textbf{Fold Structure}: For fold $k$ ($k = 0, 1$):
        \begin{itemize}
            \item Training set: tokens $[0, \text{train\_end}_k]$ where $\text{train\_end}_k = \lfloor 0.3 \times T \rfloor + k \times \text{fold\_size}$
            \item Validation set: Last 20\% of training tokens (used for hyperparameter tuning)
            \item Test set: tokens $[\text{train\_end}_k, \text{train\_end}_k + \text{fold\_size}]$
        \end{itemize}
    \item \textbf{Forecasting Task}: Given 8 input tokens (32 weeks of history), predict 1 output token (4 weeks ahead).
\end{enumerate}

This online setting differs from standard train/test splits by: (1) Respecting temporal order (no future data leakage); (2) Using expanding training windows that simulate incremental model updates; (3) Evaluating on multiple time periods to assess robustness across different epidemic phases.

\paragraph{Dataset Characteristics.}
Table~\ref{tab:downstream_datasets_detailed} provides comprehensive statistics for each downstream dataset. The diseases exhibit substantial heterogeneity in:
\begin{itemize}
    \item \textbf{Temporal coverage}: 21--122 years of observations (1888--2017)
    \item \textbf{Aggregated series length}: 1,096--6,373 weeks after aggregation
    \item \textbf{Transmission modes}: Including respiratory (12 diseases), sexually transmitted (1), fecal-oral (3), and bloodborne (1)
    \item \textbf{Basic reproduction number}: Ranging from $R_0 \approx 1$ (gonorrhea) to $R_0 \approx 12$--18 (measles, pertussis)
\end{itemize}

\begin{table}[t!]
    \centering
    \caption{Detailed statistics of downstream datasets from Project Tycho. ``Agg.\ Len.'' refers to the number of weeks after aggregating across states.}
    \label{tab:downstream_datasets_detailed}
    \scriptsize
    \setlength{\tabcolsep}{3pt}
    \begin{tabular}{lcccccr}
        \toprule
        \textbf{Disease} & \textbf{States} & \textbf{Agg.\ Len.} & \textbf{Years} & \textbf{Total Cases} & \textbf{Trans.} & \textbf{$R_0$} \\
        \midrule
        Measles & 50 & 6{,}373 & 1888--2001 & 23{,}559{,}000 & Resp. & 12--18 \\
        Diphtheria & 46 & 6{,}191 & 1888--1981 & 1{,}980{,}152 & Resp. & 1.7--4.3 \\
        \makecell[l]{Typhoid\\Fever} & 44 & 5{,}816 & 1888--1983 & 579{,}833 & Fecal & 2.8--7.0 \\
        Pertussis & 46 & 5{,}719 & 1888--2017 & 3{,}325{,}547 & Resp. & 12--17 \\
        \makecell[l]{Scarlet\\Fever} & 48 & 4{,}957 & 1888--1969 & 9{,}171{,}413 & Resp. & 0.6--2.0 \\
        Tuberculosis & 39 & 4{,}881 & 1890--2014 & 1{,}916{,}967 & Resp. & 0.24--4.3 \\
        Smallpox & 44 & 3{,}122 & 1888--1952 & 458{,}485 & Resp. & 3.5--6.0 \\
        \makecell[l]{Acute\\Poliomyelitis} & 47 & 2{,}534 & 1912--1971 & 573{,}743 & Fecal & 5--7 \\
        Mumps & 41 & 2{,}419 & 1924--2017 & 1{,}073{,}690 & Resp. & 4--7 \\
        Pneumonia & 41 & 2{,}060 & 1912--1951 & 1{,}347{,}512 & Resp. & 1.4 \\
        Influenza & 42 & 1{,}673 & 1919--1951 & 7{,}215{,}995 & Resp. & 1.2--1.6 \\
        Varicella & 30 & 1{,}571 & 1889--2017 & 1{,}973{,}108 & Resp. & 10--12 \\
        Hepatitis B & 31 & 1{,}449 & 1952--2007 & 422{,}522 & Blood & 1.0--3.3 \\
        Rubella & 7 & 1{,}427 & 1966--2017 & 191{,}091 & Resp. & 3.4--7.0 \\
        \makecell[l]{Meningococcal\\Meningitis} & 37 & 1{,}330 & 1926--1964 & 166{,}509 & Resp. & 0.6--1.6 \\
        Gonorrhea & 39 & 1{,}126 & 1972--2017 & 9{,}455{,}568 & Sexual & 1.0 \\
        Hepatitis A & 38 & 1{,}096 & 1966--2007 & 651{,}553 & Fecal & 1.1--3.5 \\
        \bottomrule
    \end{tabular}
\end{table}

\section{Implementation Details}
\label{app: implementation}

\textbf{Motivation for Our Settings.}  
Our primary focus is on pre-training epidemic forecasting models using temporal (time series) data rather than spatiotemporal data. This design choice is motivated by the following considerations:
\begin{compactenum}[\textbullet]
    \item We aim to establish the foundation for epidemic pre-training in the temporal setting, which remains unexplored — only one prior work ~\citep{kamarthi2023pems} addresses this area, and it also focuses on the temporal setting. As shown in Sections 4.2-4.5, we address critical questions around generalization and pre-training influence in the epidemic domain, which remain open even without spatial context.
    \item Temporal models are broadly applicable and more data-efficient, especially when spatial data is unavailable or unreliable. Many real-world epidemic datasets lack well-defined spatial graphs, and building them (e.g., from mobility or administrative data) is costly and complex, particularly at scale. These inconsistencies also hinder fair comparisons between temporal and spatiotemporal models.
    \item Our framework is extensible to spatiotemporal modeling. Specifically, the temporal input can be replaced with graph-structured data, and the predictor can incorporate graph-based encoders. Exploring this direction is exciting future work, but we believe temporal pre-training is a crucial first step toward that goal.
\end{compactenum}

\noindent\textbf{Model Details.} We design our model by stacking 4 layers of the CAPE encoder, each with a hidden size of 512 and 4 attention heads. For compartment representations, each is encoded with a learnable embedding of size 512 and is mapped to 6 distinct groups. The patch encoder uses a decoder-only transformer with causal masking and GELU activations. Positional encoding is applied via sinusoidal embeddings.

\noindent\textbf{Pre-training.}
We pre-train CAPE on synthetic epidemic data generated using the EpiRecipe simulation framework. The synthetic dataset consists of 5,000 validation samples and 5,000 test samples, covering diverse epidemic dynamics including SIR, SEIR, and SEIR-V models. We use a token size of 4 time steps per patch.  Training uses the AdamW optimizer with learning rate $1\times10^{-4}$, weight decay $1\times10^{-3}$, batch size 64, and CosineAnnealingWarmRestarts scheduler ($T_0=10$, $T_{\text{mult}}=2$). For the loss function, we set the $\alpha$ of the I compartment to be 1 and 0.5 for the rest of the compartments, and set the $\beta$ to be 0.5. First-stage pre-training runs for 150 epochs.
The second stage uses the first 20\% of each disease's time series in chronological order (with 10\% of this portion held for validation), ensuring strict temporal separation from downstream evaluation data. Training continues from the first-stage checkpoint using the same architecture, with AdamW optimizer (learning rate $1\times10^{-4}$, weight decay $1\times10^{-4}$), batch size 32, Huber loss ($\delta=1.0$), and 15 epochs. Only the I (Infectious) compartment receives supervision during this stage. The best checkpoint is selected based on validation loss. Both stages utilize 8 NVIDIA A100-SXM4-80GB.

\textit{PEM Pre-training.} For the PEM baseline~\citep{kamarthi2023pems}, we pre-train a PatchTST-based architecture ($d_{\text{model}}=256$, 6 layers, 8 attention heads, patch length 4, dropout 0.1) using self-supervised masked prediction. We use all 17 diseases with the first 30\% of each time series for pre-training. Three complementary masking strategies are employed: RANDMASK (random segment masking, 40\% probability), LASTMASK (masking final segments for forecasting, 30\% probability), and PEAKMASK (masking around epidemic peaks, 30\% probability). The masking ratio $\gamma$ is set to 0.3. Training uses AdamW with learning rate $1\times10^{-4}$, weight decay $1\times10^{-3}$, batch size 64, for 25 epochs.

\noindent\textbf{Downstream Evaluation.}
All models use 8 input tokens (32 weeks) to predict 1 output token (4 weeks ahead).

\paragraph{Ensemble Aggregation Strategies.}
In epidemic forecasting, underestimation poses greater risks than overestimation: underestimating an outbreak can lead to inadequate resource allocation, delayed interventions, and preventable morbidity, whereas moderate overestimation simply results in conservative preparedness. Consequently, our ensemble aggregation strategies are deliberately designed with an \emph{upward bias}, favoring predictions from more pessimistic (higher-predicting) ensemble members.

Given $M$ ensemble members with predictions $\{\hat{y}^{(m)}\}_{m=1}^{M}$, we employ 14 pooling strategies spanning conservative to aggressive estimation:

\begin{compactenum}[(1)]
    \item \textbf{percentile\_75}: 75th percentile of ensemble predictions.
    $\hat{y} = \hat{y}^{(\lceil 0.75M \rceil)}_{\text{sorted}}$
    
    \item \textbf{median\_max\_blend}: Equal blend of median and maximum.
    $\hat{y} = 0.5 \cdot \text{median}(\{\hat{y}^{(m)}\}) + 0.5 \cdot \max_m \hat{y}^{(m)}$
    
    \item \textbf{upper\_half\_mean}: Mean of predictions above the median.
    $\hat{y} = \frac{2}{M} \sum_{m: \hat{y}^{(m)} \geq \text{median}} \hat{y}^{(m)}$
    
    \item \textbf{biased\_mean\_50}: Mean with 50\% bias toward maximum.
    $\hat{y} = 0.5 \cdot \bar{y} + 0.5 \cdot \max_m \hat{y}^{(m)}$
    
    \item \textbf{optimistic\_weighted}: Weights proportional to prediction magnitude.
    $\hat{y} = \sum_m w_m \hat{y}^{(m)}$, where $w_m = \hat{y}^{(m)} / \sum_j \hat{y}^{(j)}$
    
    \item \textbf{robust\_optimal}: Maximum of (75th percentile, upper-half mean, optimistic weighted).
    $\hat{y} = \max\left(P_{75}, \bar{y}_{\text{upper}}, \hat{y}_{\text{optim}}\right)$
    
    \item \textbf{adaptive\_percentile}: Variance-adaptive percentile (70th--90th).
    $\hat{y} = P_q$ where $q = 0.7 + 0.2 \cdot \min(1, \sigma/\bar{y})$
    
    \item \textbf{trimmed\_upper}: Mean of top 70\% of predictions.
    
    \item \textbf{quantile\_blend\_60\_80}: Average of 60th and 80th percentiles.
    $\hat{y} = 0.5 \cdot P_{60} + 0.5 \cdot P_{80}$
    
    \item \textbf{shrinkage\_upper\_50}: Upper-half mean shrunk 50\% toward median.
    $\hat{y} = 0.5 \cdot \bar{y}_{\text{upper}} + 0.5 \cdot \text{median}$
    
    \item \textbf{variance\_adaptive\_blend}: Blend median/max based on coefficient of variation.
    $\hat{y} = \alpha \cdot \text{median} + (1-\alpha) \cdot \max$, where $\alpha = \max(0.3, 1 - \sigma/\bar{y})$
    
    \item \textbf{max\_plus\_std}: Maximum plus one standard deviation.
    $\hat{y} = \max_m \hat{y}^{(m)} + \sigma$
    
    \item \textbf{max\_plus\_std\_100}: Maximum plus standard deviation, scaled by 100.
    $\hat{y} = \max_m \hat{y}^{(m)} + 100 \cdot \sigma$
    
    \item \textbf{beyond\_max\_20}: Maximum extended by 20\% of ensemble range.
    $\hat{y} = \max_m \hat{y}^{(m)} + 0.2 \cdot (\max - \min)$
    \item \textbf{residual\_correction}: Observation-anchored correction: $\hat{y} = y_{\text{last\_obs}} + \alpha \cdot e^{-\text{CV}(\mathbf{r})} \cdot \text{median}(\{r^{(m)}\})$, where $r^{(m)} = \hat{y}^{(m)} - y_{\text{last\_obs}}$ and $\alpha \in \{0.2, 0.3, 0.4\}$. Corrections shrink toward the last observation when masks disagree (high CV).
\end{compactenum}

This collection spans a spectrum from moderately conservative (percentile-based) to aggressively cautious (beyond-maximum) strategies, allowing validation-based selection to identify the appropriate risk-sensitivity for each disease. During evaluation, strategy selection is performed on a held-out validation set based on MSE, and test metrics are reported for the validation-selected strategy.

\paragraph{Baseline Comparison.}
We compare against 12 baseline methods spanning statistical models (ARIMA, SIR), deep learning architectures (GRU, LSTM, CNN), specialized epidemic models (EINN, EpiDeep, PEM), and modern time series transformers (DLinear, PatchTST, NBeats, XGBoost). All baselines undergo hyperparameter tuning on the validation set within each fold, with grid search over hidden sizes $\{128, 512\}$, number of layers $\{2, 3\}$, and learning rates $\{0.001, 0.005\}$. CAPE operates in zero-shot mode with only the smoothing window size and ensemble strategy tuned on validation data.

\paragraph{Evaluation Metrics.}
Results are aggregated across all folds using mean and standard deviation. Primary metrics are Mean Squared Error (MSE) and Mean Absolute Error (MAE), computed on the final token of each prediction (4 weeks ahead). For CAPE, we additionally report uncertainty quantification via ensemble predictions using multiple random compartment masks.

\section{Representation Properties}
\label{app: rep_prop}

\begin{figure*}
    \centering
    \includegraphics[width=\linewidth]{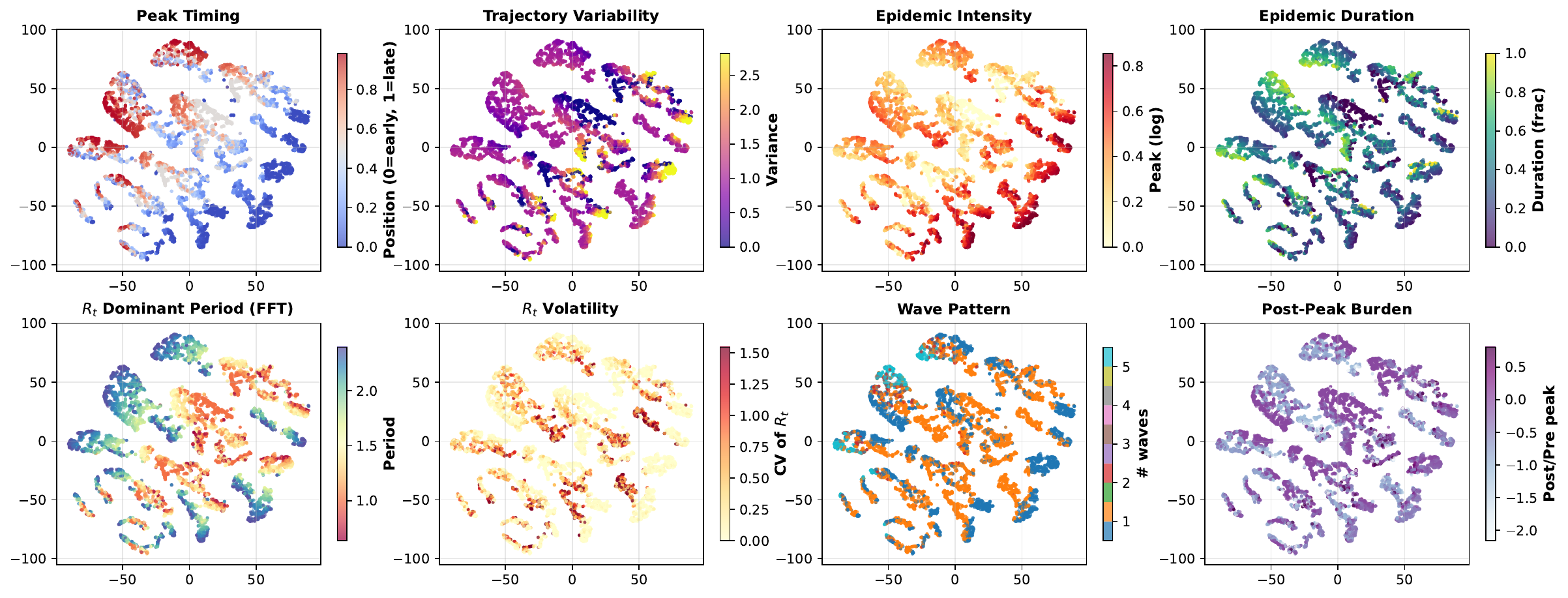}
    \caption{t-SNE visualization of CAPE encodings of synthetic samples from EpiRecipe.}
    \label{fig: full_synth_embed}
\end{figure*}

We extract the following properties from each trajectory to analyze what representations CAPE learns. Let $I(t)$ denote the infection time series of length $T$, with peak at time $t^* = \arg\max_t I(t)$ and peak value $I^* = \max_t I(t)$. The full visualization is shown in Figure~\ref{fig: full_synth_embed}.

\begin{itemize}[leftmargin=*, itemsep=2pt, topsep=2pt]
    \item \textit{Peak Timing}: Normalized position of the infection peak.
    $\tau_{\text{peak}} = t^* / T$
    
    \item \textit{Trajectory Variability}: Variance of the infection time series.
    $\sigma^2_I = \frac{1}{T} \sum_{t=1}^{T} ( I(t) - \bar{I} )^2$
    
    \item \textit{Epidemic Intensity}: Log-scaled peak height.
    $\mathcal{I} = \log_{10}(I^* + 1)$
    
    \item \textit{Epidemic Duration}: Fraction of time above 10\% of peak.
    $\mathcal{D} = \frac{1}{T} \sum_{t=1}^{T} \mathbf{1}[ I(t) > 0.1 \cdot I^* ]$
    
    \item \textit{$R_t$ Dominant Period}: Dominant periodicity from FFT of $R_t$ (log-scaled).
    $\mathcal{P} = \log_{10}( 1/f^* + 1 )$ where $f^* = \arg\max_{f > 0} | \mathcal{F}\{R_t - \bar{R}_t\}(f) |$
    
    \item \textit{$R_t$ Volatility}: Coefficient of variation of $R_t$.
    $\text{CV}_{R_t} = \sigma_{R_t} / \bar{R}_t$
    
    \item \textit{Wave Pattern}: Number of local maxima exceeding 30\% of peak height.
    $N_{\text{waves}} = | \{ t : I(t) > I(t \pm 1), \, I(t)/I^* > 0.3 \} |$
    
    \item \textit{Post-Peak Burden}: Ratio of cumulative infections after vs.\ before peak.
    $\mathcal{B} = \sum_{t > t^*} I(t) \,/\, \sum_{t \leq t^*} I(t)$
\end{itemize}

\begin{figure*}[ht]
    \centering
    \includegraphics[width=\linewidth]{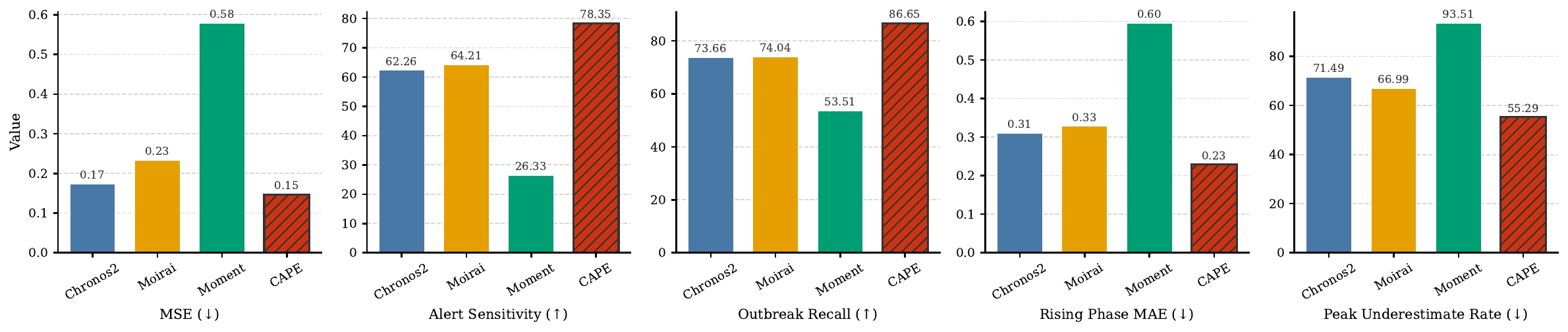}
    \vskip -1.5em
    \caption{Comparison with zero-shot foundation models across all datasets.}
    \label{fig: fm_comparison_bar}
\end{figure*}

\section{Comparison with General Time Series Foundation Models}
\label{app: radar}

We compare CAPE against three state-of-the-art time series foundation models: Chronos2, Moirai, and Moment. 
The radar plots below visualize the performance across five key epidemic forecasting metrics for each disease:

\begin{itemize}[nosep,leftmargin=*]
    \item \textbf{MSE} (Mean Squared Error, lower is better): 
    \[
    \text{MSE} = \frac{1}{N}\sum_{i=1}^{N}(\hat{y}_i - y_i)^2
    \]
    where $\hat{y}_i$ is the prediction and $y_i$ is the ground truth.
    
    

    \item \textbf{Outbreak Sensitivity} (higher is better): Measures the recall of high-value detection—when actual values exceed a threshold, how often does the model also predict above that threshold?
        \[
        \text{Outbreak Sens.} = \frac{|\{i : \hat{y}_i > \tau \land y_i > \tau\}|}{|\{i : y_i > \tau\}|} \times 100\%
        \]
        We evaluate at two threshold levels: a \emph{moderate} threshold $\tau_1 = \text{median}(y) + 0.5\sigma_y$ for general outbreak detection, and a \emph{strict} threshold $\tau_2 = \mu_y + \sigma_y$ for severe alert detection.
    
    \item \textbf{Rising Phase MAE} (lower is better): Mean Absolute Error during epidemic growth phases when values are increasing.
    \[
    \text{Rising MAE} = \frac{1}{|R|}\sum_{i \in R}|\hat{y}_i - y_i|, \quad R = \{i : y_i - x_i^{\text{last}} > 0.1\}
    \]
    where $x_i^{\text{last}}$ is the last observed value before prediction and $R$ is the set of rising-phase samples.
    
    \item \textbf{Peak Underestimate Rate} (lower is better): Among peak values (top 10\%), how often does the model underestimate?
    \[
    \text{Peak Underest.} = \frac{|\{i : \hat{y}_i < y_i \land y_i \geq P_{90}(y)\}|}{|\{i : y_i \geq P_{90}(y)\}|} \times 100\%
    \]
    where $P_{90}(y)$ is the 90th percentile of target values.
\end{itemize}

\noindent\textbf{Normalization for Radar Plots:} For each disease, metric values are normalized across the four models using min-max scaling to a 0--100 range. For ``lower is better'' metrics (MSE, Rising Phase MAE, Peak Underestimate Rate), the scale is inverted so that values closer to the outer edge (100) consistently indicate better performance:
\[
\text{score} = \begin{cases}
\dfrac{v - v_{\min}}{v_{\max} - v_{\min}} \times 100 & \text{if higher is better} \\[1.5ex]
100 - \dfrac{v - v_{\min}}{v_{\max} - v_{\min}} \times 100 & \text{if lower is better}
\end{cases}
\]
where $v$ is the raw metric value and $v_{\min}$, $v_{\max}$ are the minimum and maximum values across the four models for that metric.

\begin{figure*}[htbp]
    \centering
    \begin{subfigure}[b]{0.32\textwidth}
        \centering
        \includegraphics[width=\textwidth]{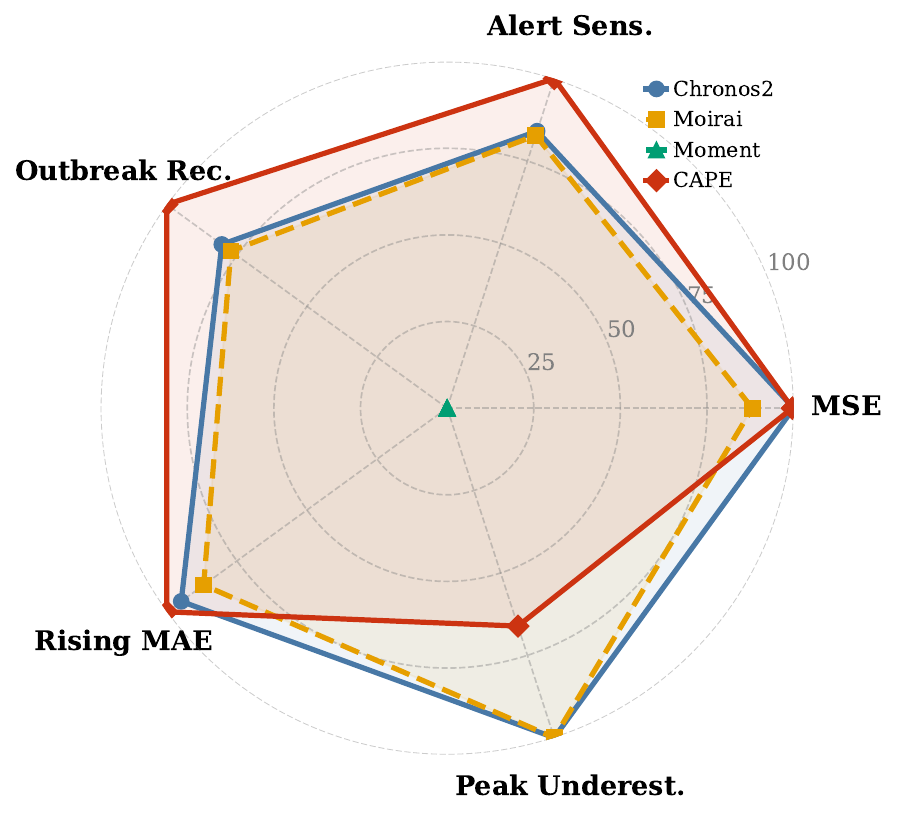}
        \caption{Acute Poliomyelitis}
    \end{subfigure}\hfill
    \begin{subfigure}[b]{0.32\textwidth}
        \centering
        \includegraphics[width=\textwidth]{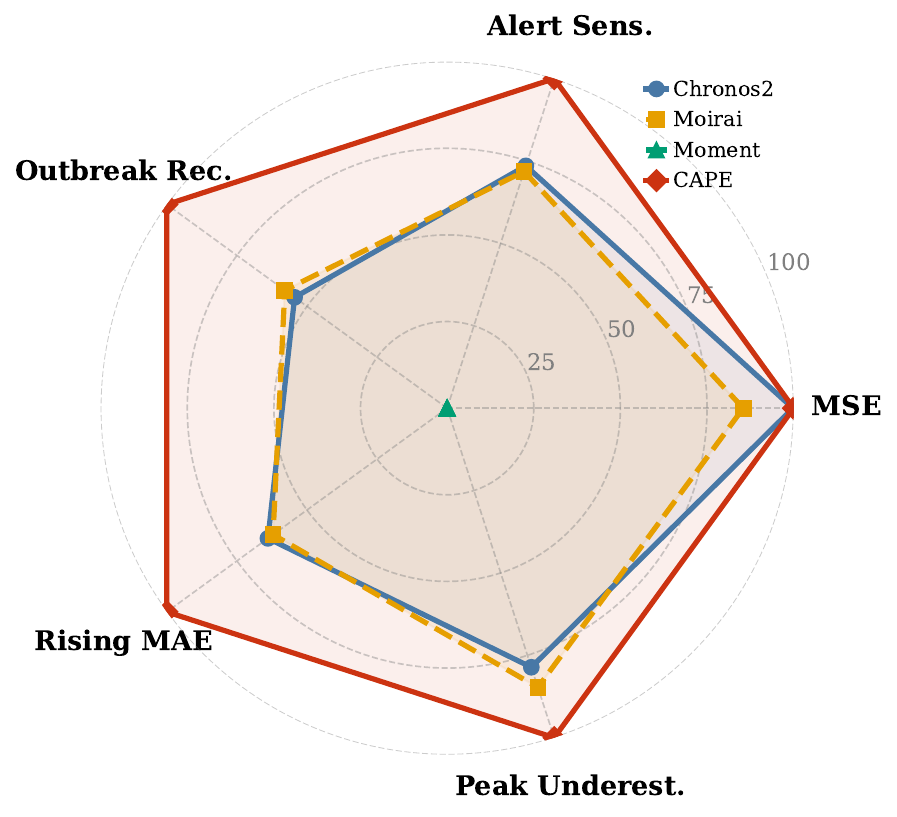}
        \caption{Diphtheria}
    \end{subfigure}\hfill
    \begin{subfigure}[b]{0.32\textwidth}
        \centering
        \includegraphics[width=\textwidth]{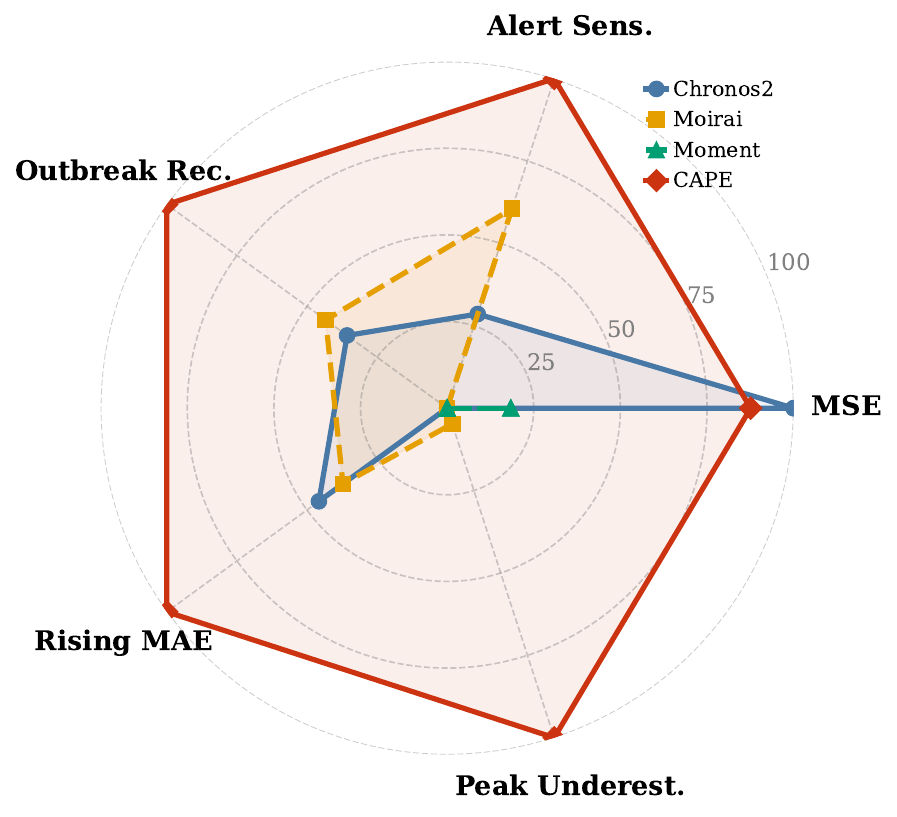}
        \caption{Gonorrhea}
    \end{subfigure}\\[0.5ex]
    \begin{subfigure}[b]{0.32\textwidth}
        \centering
        \includegraphics[width=\textwidth]{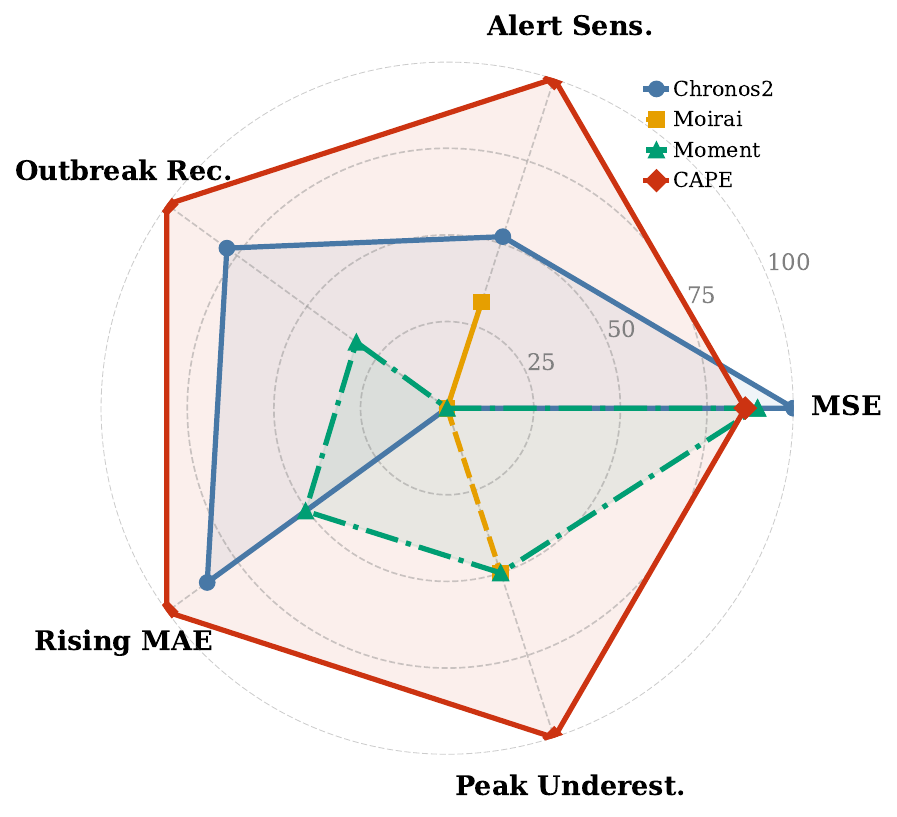}
        \caption{Hepatitis A}
    \end{subfigure}\hfill
    \begin{subfigure}[b]{0.32\textwidth}
        \centering
        \includegraphics[width=\textwidth]{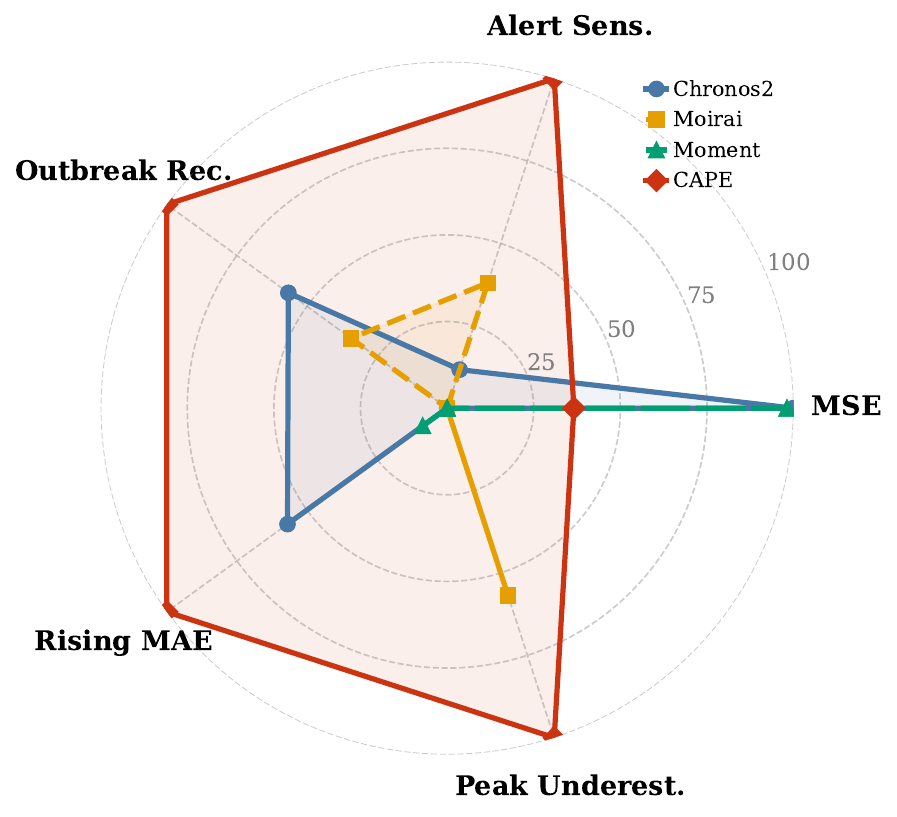}
        \caption{Hepatitis B}
    \end{subfigure}\hfill
    \begin{subfigure}[b]{0.32\textwidth}
        \centering
        \includegraphics[width=\textwidth]{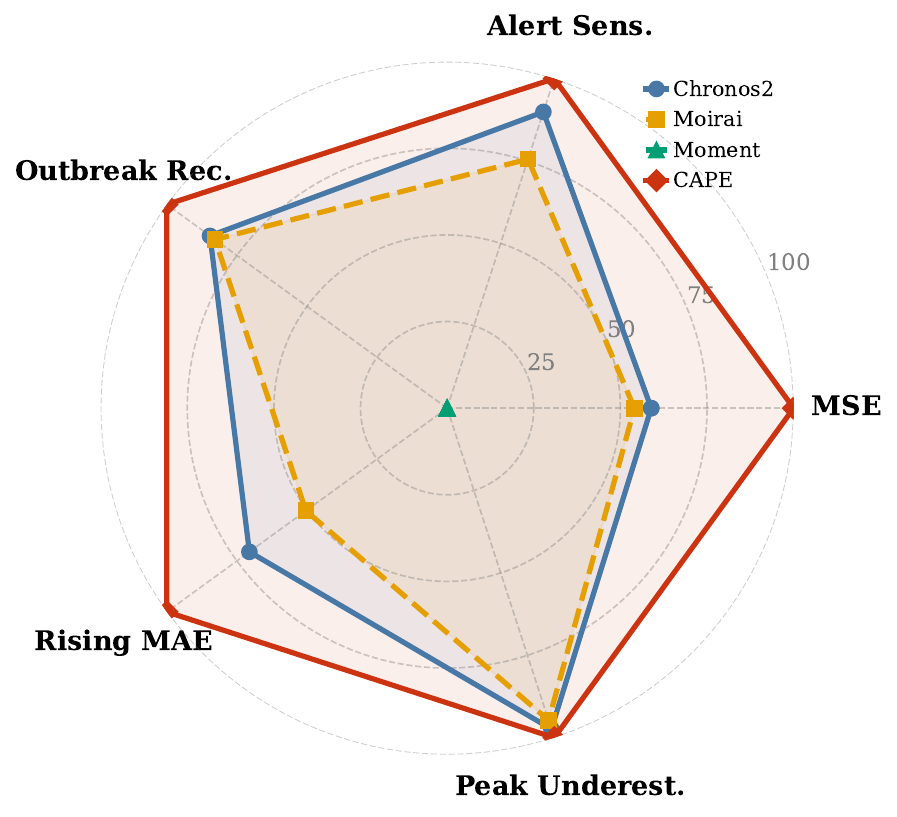}
        \caption{Influenza}
    \end{subfigure}\\[0.5ex]
    \begin{subfigure}[b]{0.32\textwidth}
        \centering
        \includegraphics[width=\textwidth]{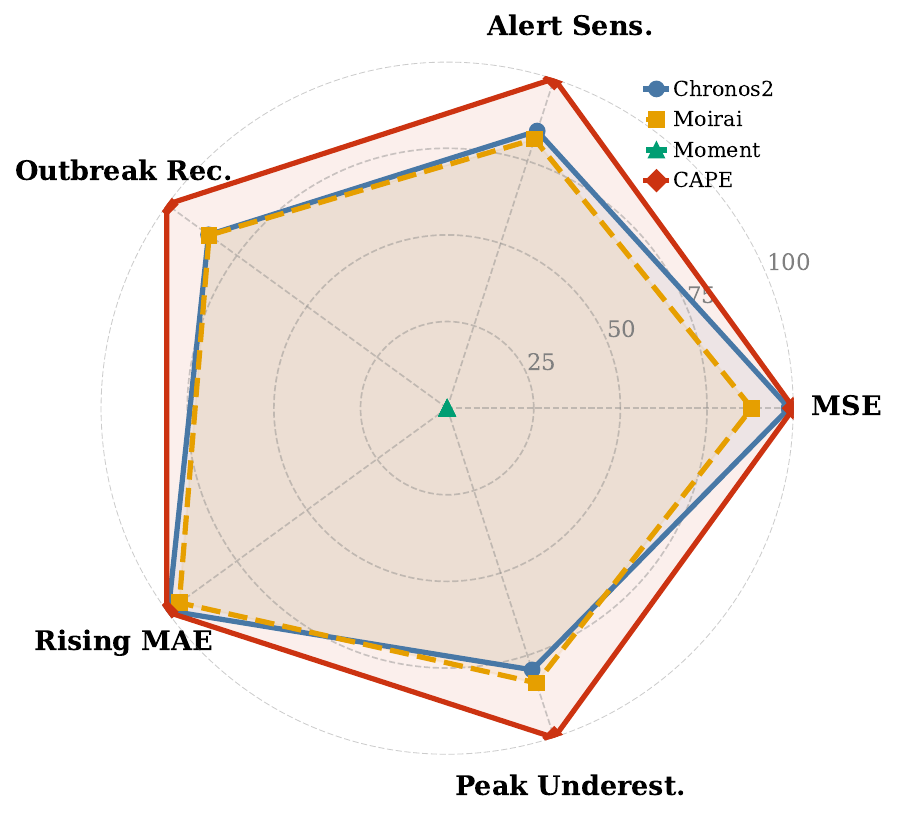}
        \caption{Measles}
    \end{subfigure}\hfill
    \begin{subfigure}[b]{0.32\textwidth}
        \centering
        \includegraphics[width=\textwidth]{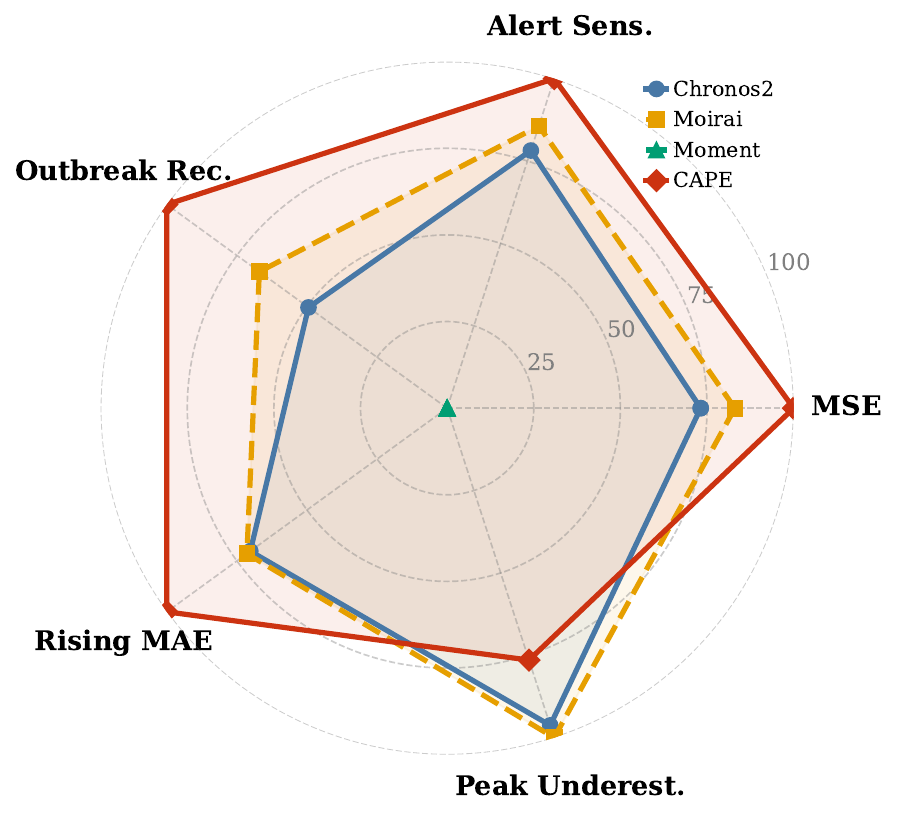}
        \caption{Meningococcal Meningitis}
    \end{subfigure}\hfill
    \begin{subfigure}[b]{0.32\textwidth}
        \centering
        \includegraphics[width=\textwidth]{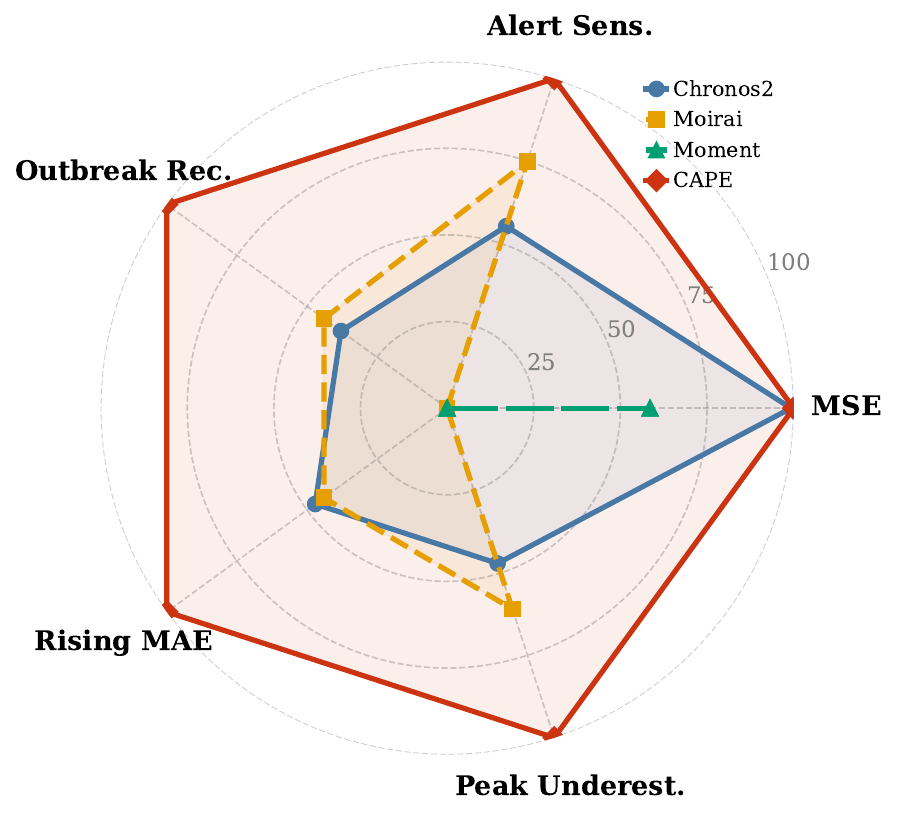}
        \caption{Mumps}
    \end{subfigure}
    \caption{Performance comparison across epidemic forecasting metrics (Part 1 of 2). Each radar plot shows five normalized metrics. Values closer to the outer edge indicate better performance.}
    \label{fig:radar-comparison-1}
\end{figure*}

\begin{figure*}[htbp]
    \centering
    \begin{subfigure}[b]{0.32\textwidth}
        \centering
        \includegraphics[width=\textwidth]{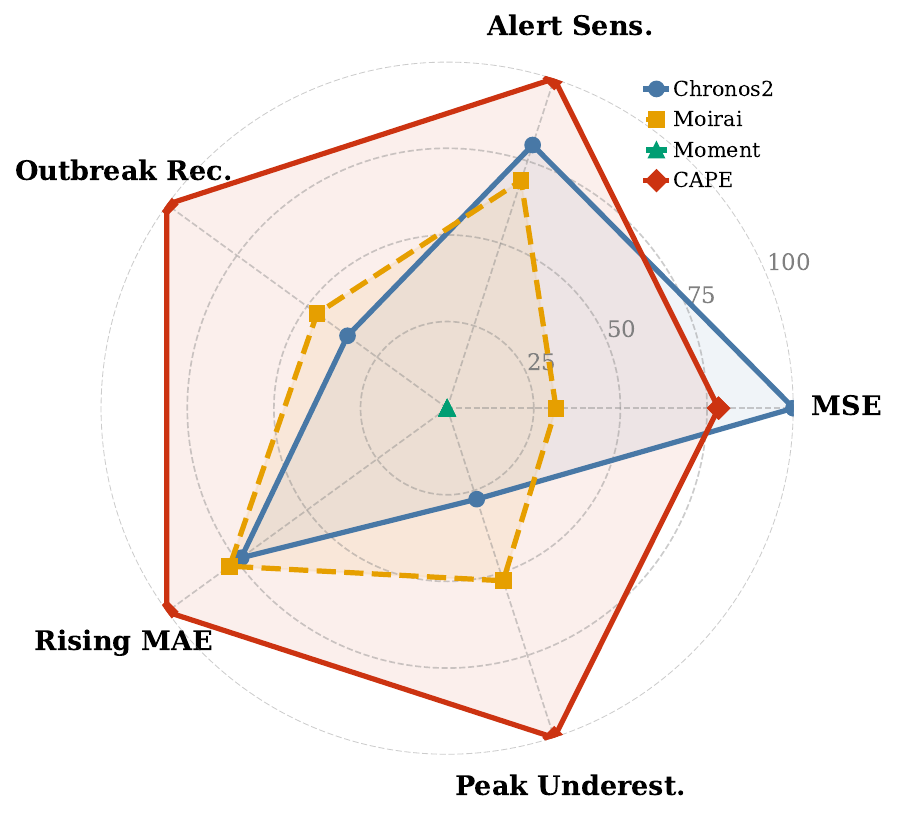}
        \caption{Pertussis}
    \end{subfigure}\hfill
    \begin{subfigure}[b]{0.32\textwidth}
        \centering
        \includegraphics[width=\textwidth]{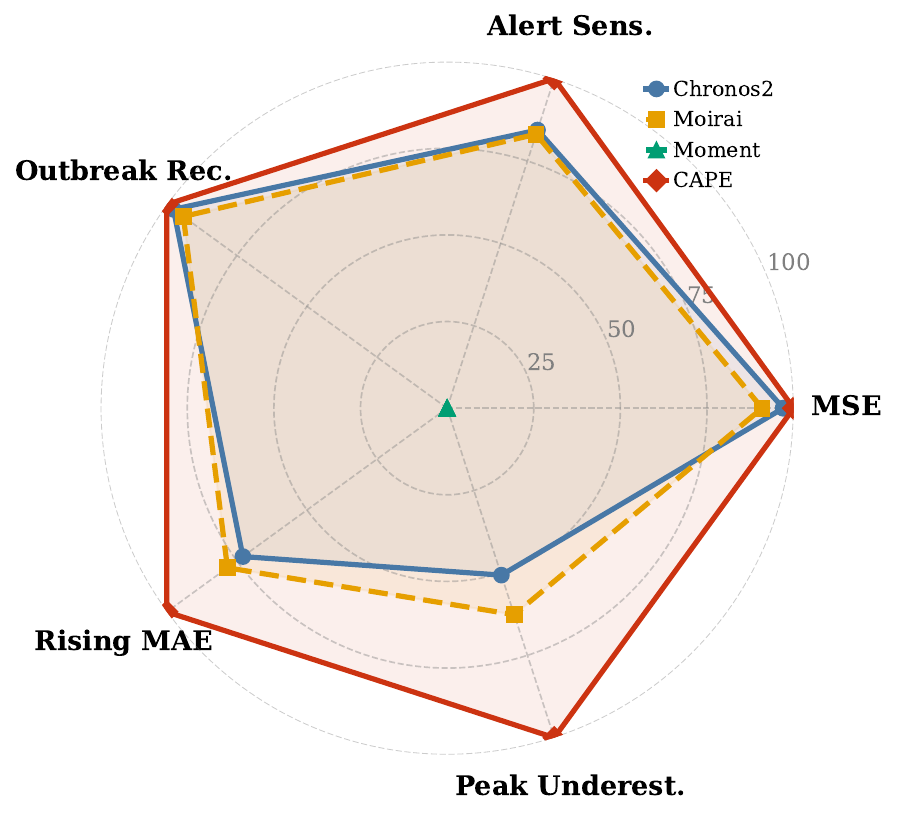}
        \caption{Pneumonia}
    \end{subfigure}\hfill
    \begin{subfigure}[b]{0.32\textwidth}
        \centering
        \includegraphics[width=\textwidth]{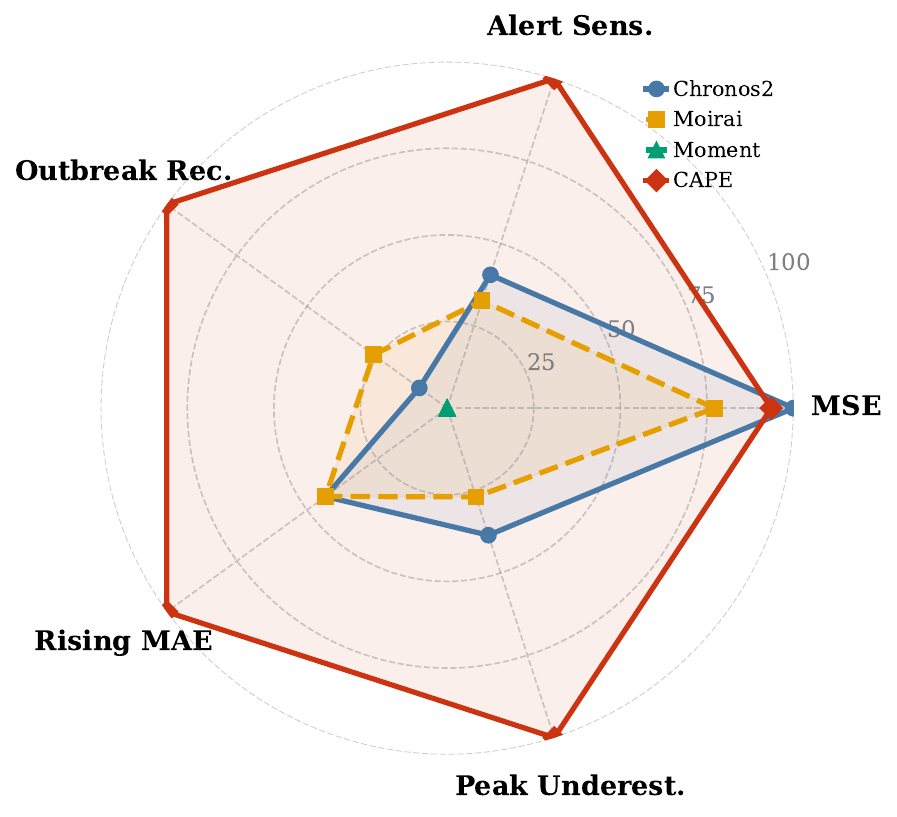}
        \caption{Rubella}
    \end{subfigure}\\[0.5ex]
    \begin{subfigure}[b]{0.32\textwidth}
        \centering
        \includegraphics[width=\textwidth]{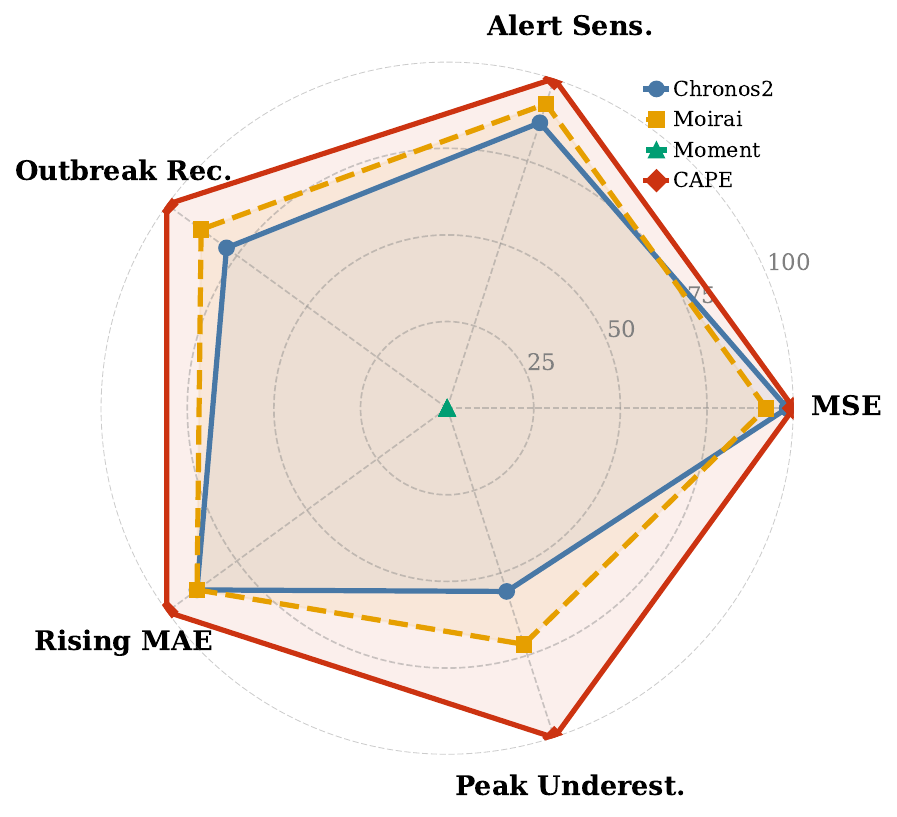}
        \caption{Scarlet Fever}
    \end{subfigure}\hfill
    \begin{subfigure}[b]{0.32\textwidth}
        \centering
        \includegraphics[width=\textwidth]{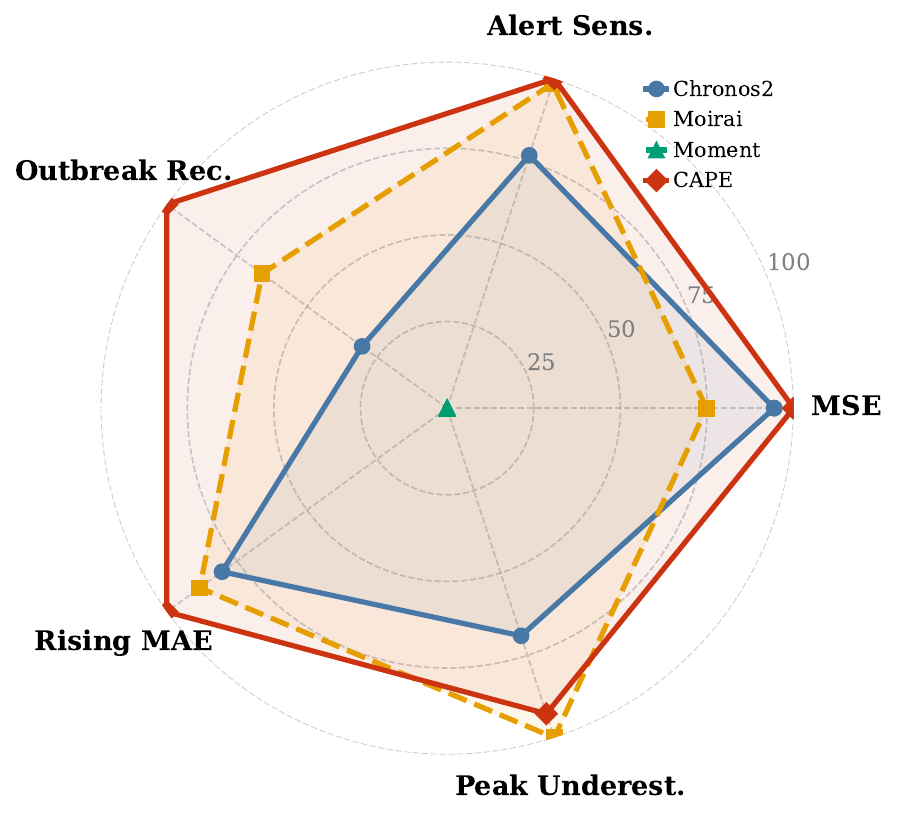}
        \caption{Smallpox}
    \end{subfigure}\hfill
    \begin{subfigure}[b]{0.32\textwidth}
        \centering
        \includegraphics[width=\textwidth]{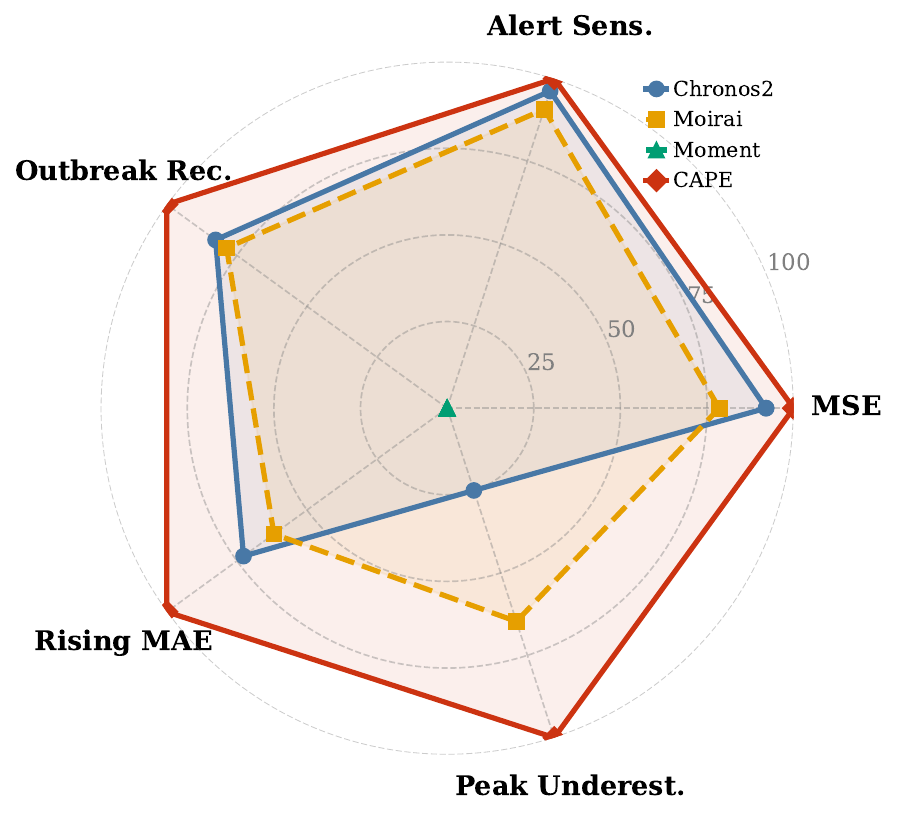}
        \caption{Tuberculosis}
    \end{subfigure}\\[0.5ex]
    \begin{subfigure}[b]{0.32\textwidth}
        \centering
        \includegraphics[width=\textwidth]{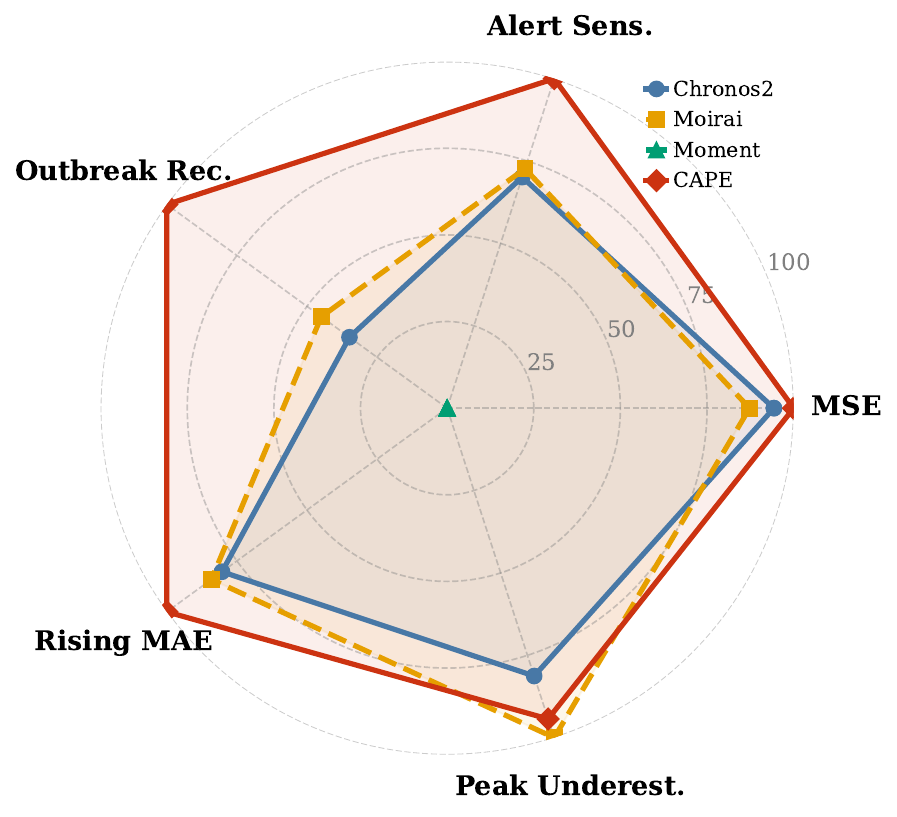}
        \caption{Typhoid Fever}
    \end{subfigure}\hfill
    \begin{subfigure}[b]{0.32\textwidth}
        \centering
        \includegraphics[width=\textwidth]{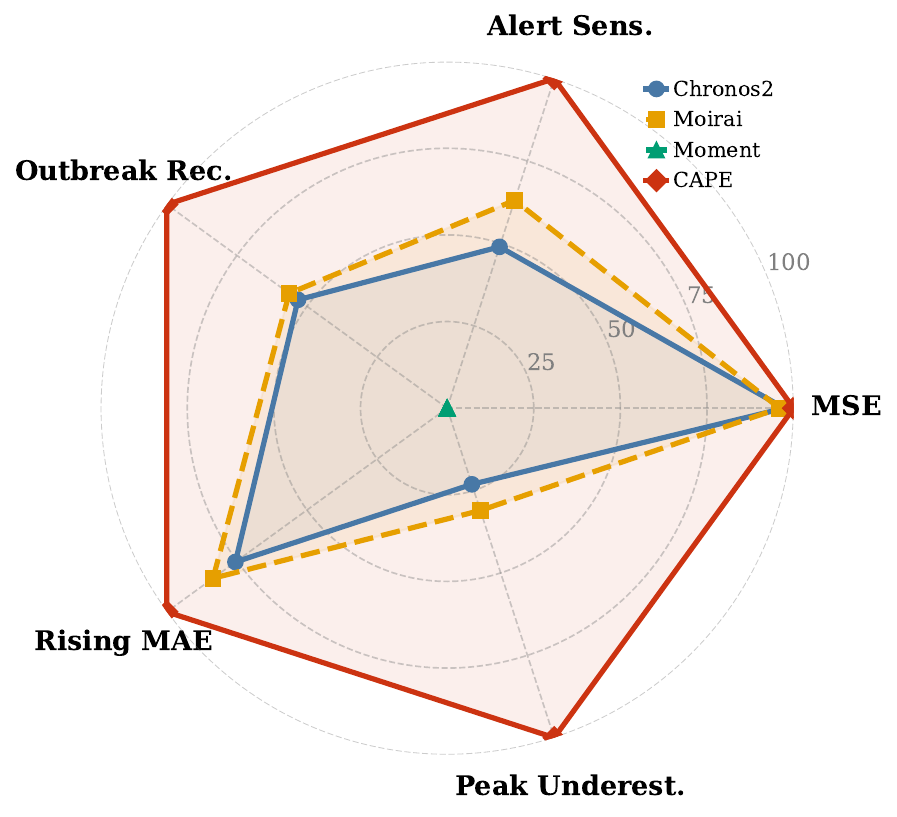}
        \caption{Varicella}
    \end{subfigure}\hfill
    \begin{subfigure}[b]{0.32\textwidth}
        \centering
        \includegraphics[width=\textwidth]{figures/overall_radar_alert_sensitivity.pdf}
        \caption{\textbf{Overall Average}}
    \end{subfigure}
    \caption{Performance comparison across epidemic forecasting metrics (Part 2 of 2). The ``Overall Average'' shows mean performance across all 17 diseases.}
    \label{fig:radar-comparison-2}
\end{figure*}

\end{document}